\documentclass[journal, onecolumn]{IEEEtran}
\IEEEoverridecommandlockouts
\usepackage{amsmath,amssymb,amsfonts}
\usepackage{algorithmic}
\usepackage{graphicx}
\usepackage{textcomp}
\usepackage{xcolor}
\usepackage{booktabs}
\usepackage{multirow}
\usepackage[caption=false,font=footnotesize]{subfig}
\usepackage{hyperref}
\usepackage[numbers, sort&compress]{natbib}
\begin{document}

\title{Influence-Based Reward Modulation for Implicit Communication in Human-Robot Interaction\thanks{© 2026 IEEE. Personal use of this material is permitted. Permission from IEEE must be obtained for all other uses.

This is the accepted version of the paper: Haoyang Jiang, Elizabeth A. Croft, and Michael G. Burke. 2026. Influence-Based Reward Modulation for Implicit Communication in Human-Robot Interaction. J. Hum.-Robot Interact. Just Accepted (March 2026). https://doi.org/10.1145/3800955}}

\author{
    Haoyang Jiang\textsuperscript{1},  
    Elizabeth A. Croft\textsuperscript{2},  
    Michael G. Burke\textsuperscript{1} \\  
    \textsuperscript{1}Monash University, Melbourne, Australia \\  
    \textsuperscript{2}University of Victoria, Victoria, Canada \\  
    haoyang.jiang@monash.edu, ecroft@uvic.ca, michael.g.burke@monash.edu  
}

\maketitle

\begin{abstract}
	Communication is essential for successful interaction. In human-robot interaction, implicit communication holds the potential to enhance robots' understanding of human needs, emotions, and intentions. This paper introduces a method to foster implicit communication in HRI without explicitly modelling human intentions or relying on pre-existing knowledge. Leveraging Transfer Entropy, we modulate influence between agents in social interactions in scenarios involving either collaboration or competition. By integrating influence into agents' rewards within a partially observable Markov decision process, we demonstrate that boosting influence enhances collaboration and interaction, while resisting influence promotes social independence and diminishes performance in certain scenarios. Our findings are validated through simulations and real-world experiments with human participants in social navigation and autonomous driving settings.
\end{abstract}

\section{Introduction}
\label{sec:introduction}

It is widely recognised that communication is key to successful interaction. Humans communicate with each other through both explicit (direct, deliberate communication over an established channel with clear intent to reach a defined recipient \cite{Naomi2018}) and implicit channels. Implicit communication is a subtle, indirect mode of conveying information, often relying on context, nonverbal cues, and shared understanding between communicators to convey meaning without explicit verbalisation \cite{Breazeal2005}. Implicit communication is particularly important for human-robot interaction as it enhances a robot's ability to proactively understand and respond to human needs, emotions, and intentions, thereby facilitating more natural and effective communication and collaboration between humans and robots. To date, most current human-robot interaction (HRI) studies focusing on implicit communication explicitly model the intention of human participants \cite{che2020} \cite{Sadigh2016PlanningFA}, or rely on existing intention knowledge \cite{Li2014}. This information is challenging to obtain in general settings.

\begin{figure}[t]
	\centering
	\includegraphics[width=0.6\linewidth]{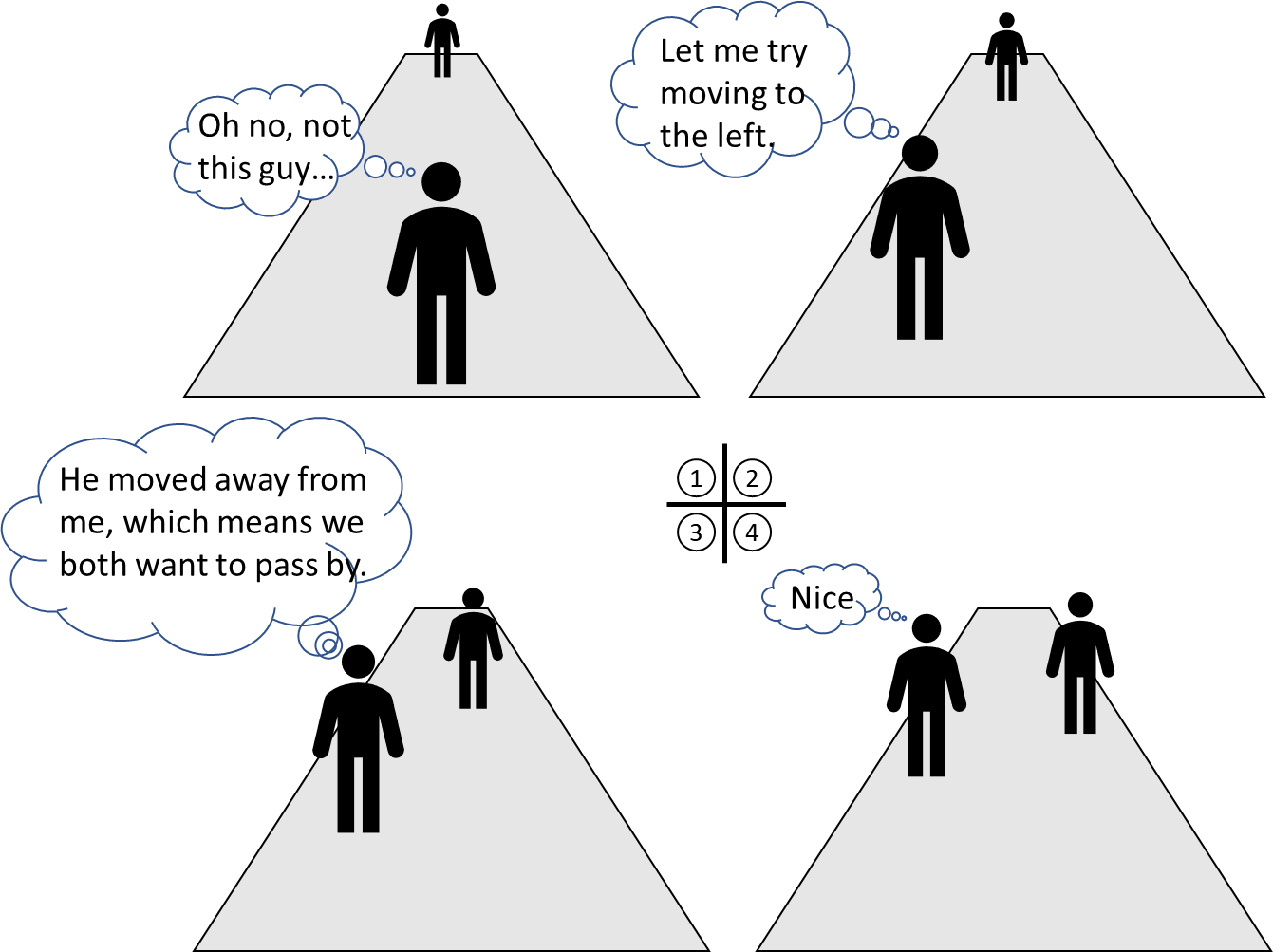}
	\caption{Humans rely on implicit communication and information exchange to negotiate complex social settings. This work introduces a reward augmentation framework to emulate this implicit communication behaviour. We show that by optimising robot policies to increase information flow, we can influence human behaviour and improve human-robot collaboration.}
	\label{fig:comic}
\end{figure}

This paper proposes a method to facilitate implicit communication without the need to explicitly model human participants or rely on pre-existing knowledge. Our approach conceptualises communication as the degree of influence agents have on one another, employing information-theoretic techniques. Specifically, we use Transfer Entropy (TE) \cite{Schreiber2000}, a measure of directed information transfer, to modulate influence in social interactions. We initially illustrate our method firstly using a social navigation setting \cite{Mavrogiannis2023}, targeting scenarios where the objectives of individual agents may result in a need for either collaboration or competition. In particular, we demonstrate the proposed method in a \textit{corridor dilemma} setting, a popular scenario in HRI \cite{Senft2020} \cite{Dondrup2014}. For our experiments, we define this as a scenario where individuals encounter each other in a narrow passage and where a given agent may wish to meet or avoid another agent. Figure \ref{fig:comic} illustrates implicit coordination between two pedestrians during such a scenario. As they approach each other, one person initiates a sidestep to the left, providing a subtle social cue. The other interprets this movement as an intent to pass, responds accordingly, and both adjust their paths without verbal communication. The successful outcome highlights how mutual understanding can emerge through minimal, yet meaningful, non-verbal signals. This scenario involves implicit information exchange, where participants attempt to discern each other's intentions without verbal communication. By observing and reacting to subtle cues, such as changes in trajectories, participants estimate each other's objectives to inform their own actions. Interactions of this form can be complex, especially when both participants act simultaneously, leading to a state of "resonance" that dissipates as mutual understanding is achieved.

To better equip robots with the ability to handle this type of implicit social interaction setting, we treat interaction as a partially observable Markov decision process (POMDP) and propose a novel approach that augments the rewards to agents to include the TE that boosts or resists influence between agents. We first evaluate the proposed framework in a controlled discretised environment involving only two agents. We find that boosting influence enhances the other participant's ability (robot or human) to collaborate or compete with the ego-agent, while resisting influence mainly diminishes their ability to collaborate. These findings hold both in self-play \cite{Silver2017MasteringCA} simulations and in virtual human-agent experiments. Real-world human-robot experiments with physical robots also show that boosting influence improved the human's collaboration performance, but reduced their competition performance. 

While the controlled discretised setting allows for clear analysis of influence modulation, real-world HRI often involve continuous state spaces, more complex dynamics, and interactions with multiple agents. To examine whether the proposed framework generalises beyond simple, two-agent scenarios, we extend it to deep reinforcement learning and multi-agent settings, and evaluate its performance in the Highway environment from the \textit{highway-env} collection \cite{highway-env}. This task features a continuous state space, more complex dynamics, and one-to-many interactions. The results show that boosting influence promotes interaction, leading to more assertive driving behaviours, whereas suppressing influence results in more conservative driving behaviours. This experiment further demonstrates the applicability and generality of the proposed framework.

Overall, our work makes the following contributions and findings:
\begin{itemize}
	\item We propose a novel framework for enhancing implicit communication in social HRI through influence modulation, without requiring explicit modelling or prior knowledge of humans.
	\item We evaluate the framework in both simulation and virtual human-agent interaction corridor dilemma environments, finding that promoting information transfer promotes collaboration and benefits the human in competitive scenarios.
	\item We show that the general trend holds in real-world human-robot corridor dilemma experiments, with differences observed.
	\item We evaluate the framework in the simulated multi-agent Highway task, finding that promoting information transfer enhances interaction and leads to more assertive behaviours, while suppressing information transfer results in more conservative behaviours.
\end{itemize}


\section{Related Work}
\label{sec:related_work}

\textbf{Implicit communication}
has attracted significant attention in the field of robotics. \citet{che2020} present a planning framework that integrates implicit and explicit communication during mobile robot navigation, using a human behaviour model and inverse reinforcement learning to generate communicative actions that enhance robot transparency and efficiency. \citet{songpo2014} \cite{songpo2017} introduce a novel interaction framework that enables communication between humans and assistive devices, leveraging gaze movements to infer human intent. \citet{liang2019} investigate the role of implicature, a form of implicit communication, in HRI using the cooperative card game Hanabi, to demonstrate that human players paired with an implicature AI are significantly more likely to perceive their partner as human. \citet{Tian2018LearningTC} presented a framework for implicit communication based on belief models that represent how other agents infer the ego agent’s intentions. These belief models must be trained separately from the acting policies, and distinct models are required when agents function differently within the environment, which requires prior knowledge of the tasks. More recent work \cite{Han2025} proposed constructing implicit communication channels using scouting actions, which are a subset of non-intrusive actions that implicitly convey information among agents. However, these scouting actions must still be manually defined by the users, requiring prior knowledge. Most implicit communication studies typically involve explicit modelling of other participants or rely on pre-existing social knowledge \cite{Li2014}. In contrast, our approach does not require any additional human model, and does not require pre-existing social knowledge.

Implicit communication is closely related to intent communication, which aims to allow observers to easily deduce the intent behind generated  behaviours \cite{busch_learning_2017} \cite{dragan_legibility_2013}. Motions conveying intent have been shown to enhance perceived safety in virtual human-robot path-crossing tasks \cite{lichtenthaler_influence_2012}. In reinforcement learning settings, robustness, efficiency and energy reward terms have been shown to enhance the ability of humans to interpret a robot's intent \cite{busch_learning_2017}. Research focusing on projected visual legibility cues \cite{hetherington_hey_2021} has indicated that projected arrows generally offer greater interpretability than flashing lights in navigation settings. Notably, the means through which people convey and comprehend intent vary, and there exists no universally applicable method for assessing intent communication. By viewing communication as the exchange of information between agents, our TE-based framework holds the potential to adapt to various forms of communication, offering a versatile method to shape communication dynamics. Importantly, unlike many of the uni-directional intent communication approaches above, our framework captures the multi-directional nature of communication.

\textbf{Transfer Entropy (TE)} is a mathematical concept introduced by \citet{Schreiber2000} that quantifies the directional flow of information between two random processes. Formally, TE is defined as the difference between the conditional entropies of one process given the past of another and the past of itself. This is expressed mathematically as
\begin{equation}
	\begin{aligned}
		TE(X \rightarrow Y) &= H(Y_t | Y_{t-1}, Y_{t-2}, \ldots)  - H(Y_t | Y_{t-1}, Y_{t-2}, \ldots, X_{t-1}, X_{t-2}, \ldots),
	\end{aligned}
\end{equation}
where \(H\) denotes entropy, and \(X\) and \(Y\) are the two stochastic processes under consideration. By measuring how much uncertainty about the future state of one process is reduced when the past of another process is known, TE provides a rigorous framework for assessing information transfer. This ability to capture not just correlations but also the causal influence between processes makes TE particularly useful for analysing complex interactions in diverse fields such as neuroscience, economics, and human behaviour, where understanding the directionality of information flow is crucial. TE is most prevalent in economics literature \cite{baek_transfer_2005,he_comparison_2017}, but has also been used to analyse animal-animal or animal-robot interactions  \cite{shaffer_transfer_2020}\cite{porfiri_inferring_2018},  joint attention \cite{sumioka_causality_2007} and to model pedestrian evacuation \cite{xie_detecting_2022}. \citet{berger_transfer_2014} apply TE to detect human-to-robot perturbations using low-cost sensors. TE has also been used to identify arbitrary cues from raw social interaction data between humans \cite{jiang2024}. These works successfully quantify information transfer or the relationship between sources of information, but tend to focus on specific features or aspects of interest. Our work seeks to extend the use of TE to provide a more general approach aimed at enhancing information transfer in decision-making settings, leading to improved human-robot collaboration. Our core assumption is that greater transparency or information transfer reduces information asymmetry \cite{Wilkinson2005}, thereby enhancing collaboration potential. In this paper we leverage TE to control information asymmetry, and tip the balance of power in favour of or against opponents.

\textbf{Social learning:}
Mutual information (MI) is an information measure that captures the shared information between random variables, which has been widely used for social learning. \citet{klyubin_empowerment_2005} proposed the concept of empowerment for intrinsic motivation for reinforcement learning, which measures the maximum MI (channel capacity) between an agent's actuators and their sensors. This concept was expanded by Mohamed and Rezende with a lower complexity maximisation approach to MI \cite{mohamed_variational_2015}. Jaques et al. use MI as a social influence reward for multi-agent deep reinforcement learning in Sequential Social Dilemmas (SSDs) \cite{leibo_multi-agent_2017} to encourage collaborations between agents \cite{jaques_social_2019}. While MI measures the shared information between random variables, TE measures the time-asymmetric information transfer. Unlike MI, the asymmetric nature of TE enables us to capture the directionality of information flow, which is advantageous for social communication. By leveraging TE, our goal is to quantify and modulate social information transfer.


\section{Methodology}
\label{sec:methodology}

We consider an environment with multiple agents, each possessing distinct states, actions, and rewards. These rewards can be complementary (collaborative), competitive, or unrelated. The ego-agent aims to maximise its own rewards, ideally without negatively impacting other agents. In an HRI context, we posit that the robot should exhibit behaviour perceived as altruism. A fair outcome involves promoting collaboration, yielding to humans in competitive scenarios, and allowing independent agents to achieve their objectives if there is no conflict of interest. Within this environment, the ego-agent has complete access to its own state and can observe other agents, but may lack access to their decision-making processes or complete states. Consequently, we model interaction as a partially observable Markov decision process (POMDP), where states represent system situations and actions determine transitions between these states. Observations derived from states guide decision-making despite incomplete information.

Our formulation is closely aligned with the Active Inference framework, which models perception and action as processes of inference under partial observability and uncertainty \cite{friston_active_2022}. While Active Inference typically focuses on minimising uncertainty through belief updates, our approach instead promotes information transfer by rewarding influence between agents. In this sense, influence modulation serves as a complementary mechanism that facilitates mutual adaptation and predictability through interaction, without requiring explicit inference over others’ internal states.

The core concept behind the proposed method involves leveraging influence as an additional reward to enhance implicit communication between agents. This augmentation aims to accelerate learning and ultimately improve the collaboration performance for all agents or the competitive capability of the other agents. Below, we assume two agents, labelled as $P1$ and $P2$ with $P1$ the ego-agent. By computing the TE from the historical actions of $P2$ (other agent) to the current action of $P1$ (ego-agent), P1 can better respond to P2 without an explicit model. This TE is computed as $TE_{\mathbf{a}_{2,t}^{(n)}\rightarrow a_{1,t}}$, where $a_{1,t}$ is ego-agent's action at time $t$, $\mathbf{a}_{2,t}^{(n)}=(a_{2,t-1},a_{2,t-2},...,a_{2,t-n})$ comprises historical actions of $P2$, and $n$ is the history length. We augment the TE into the reward function, 
\begin{equation}
	\label{add_te2reward}
	\begin{aligned}
		Reward_{a_{1,t}} = \phi TE_{\mathbf{a}_{2,t}^{(n)}\rightarrow a_{1,t}} + r.
	\end{aligned}
\end{equation}
Here, $r$ is the reward related to the agent's objectives and $\phi$ is a scaling factor used to adjust the strength of the TE reward. Importantly, augmenting the reward with TE encourages the ego-agent to take actions that promote influence of past actions of the other agent on the current action of the ego-agent, making it more legible \cite{dragan_legibility_2013} to the other agent.  Legibility increases because the ego-agent's response to the other agent's actions is clearer, allowing the ego agent to affect other agents without requiring access to their behaviour models. More specifically, by promoting influence, we encourage the trained agent to choose actions that lead to states where it's behaviour in response to other agent's behaviour is more legible to the other agent.

\subsection{Measuring influence}
\label{subsec:transfer_entropy}

We manipulate the influence using the Transfer Entropy, $T_{Y\rightarrow X}$, between the time series of state and action pairs. Transfer entropy is closely related to \textit{Wiener-Granger causality} \cite{granger_investigating_1969} and computes the conditional mutual information between two variables $X_t$ and $Y_t$ \cite{bossomaier_transfer_2016}, 
\begin{equation}
	\label{TE}
	\begin{aligned}
		T_{Y\rightarrow X}^{(k,l)}(t) & =I(X_t:\mathbf{Y}_{t}^{(l)}|\mathbf{X}_{t}^{(k)})
		= H(X_t|\mathbf{X}_{t}^{(k)})-H(X_t|\mathbf{X}_{t}^{(k)},\mathbf{Y}_{t}^{(l)}).
	\end{aligned}
\end{equation}
Here, $I$ denotes mutual information, while $H$ represents Shannon's entropy \cite{shannon_mathematical_1948}. This equation measures the directed transfer of information from $Y$ to $X$, with $t$ indicating time. The terms $\mathbf{Y}{t}^{(l)}$ and $\mathbf{X}{t}^{(k)}$ denote sequences of past observations for $Y$ and $X$, respectively, with lengths $k$ and $l$,
\begin{equation}
	\label{define_embedding}
	\begin{aligned}
		& \mathbf{X}_{t}^{(k)} = (x(t-\delta), x(t-2\delta), ..., x(t-(k-1)\delta))\\
		& \mathbf{Y}_{t}^{(l)} = (y(t-\delta), y(t-2\delta), ..., y(t-(l-1)\delta)).
	\end{aligned}
\end{equation}
Here $\delta$ is the unit time step. In literature, $X$ is referred to as the `target' and $Y$ as the `source'. Transfer entropy can be intuitively understood as the decrease in uncertainty within a state $X$, predicted solely based on its own past, upon the introduction of an additional information source $Y$ \cite{bossomaier_transfer_2016}. 

In order to compute TE, we calculate two entropy terms, the first assumes that there is influence from past actions of the other agent $P2$ to the current action of the ego-agent $P1$ (superscript $^{+}$), while the second assumes no influence (superscript $^{-}$). We denote \(\mathbf{s}_{1,t}^{(n)}=(s_{1,t}, s_{1,t-1}, s_{1,t-2}, ..., s_{1,t-n})\) and \(\mathbf{o}_{2,t}^{(n)}=(o_{2,t}, o_{2,t-1}, o_{2,t-2}, ..., o_{2,t-n})\), where $s_{1,t}, s_{1,t-1}, ..., s_{1,t-n}$ and $o_{2,t}, o_{2,t-1}, ..., o_{2,t-n}$ comprise the states of the ego-agent $P1$ and observations of measurable state information from agent $P2$ at time $t, t-1, ..., t-n$ respectively. We then compute probability distributions over the actions (policies) for each of these scenarios:
\begin{equation}
	\label{action_distribution}
	\begin{aligned}
		P_{a_{1,t}}^{+} &= P(a_{1,t} | \mathbf{s}_{1,t}^{(n)}, \mathbf{o}_{2,t}^{(n)}), \\
		P_{a_{1,t}}^{-} &= P(a_{1,t} | \mathbf{s}_{1,t}^{(n)})
	\end{aligned}
\end{equation}
and calculate Shannon's entropy. Here we show the formulation for discrete Shannon's entropy, but Shannon's differential entropy should be used if the action space is continuous. We can now calculate the TE, or information flowing from past actions of $P2$ to the current action of $P1$.
\begin{equation}
	\label{get_te}
	\begin{aligned}
		&TE_{\mathbf{a}_{2,t}^{(n)}\rightarrow a_{1,t}} = TE_{f(\mathbf{o}_{2,t}^{(n)})\rightarrow a_{1,t}} 
		= H(P_{a_{1,t}}^{-}) - H(P_{a_{1,t}}^{+}) \\
		&= \bigg[-\sum_{p \in P_{a_{1,t}}^{-}} 
		p(a_{1,t} | \mathbf{s}_{1,t}^{(n)}) 
		\log_2 p(a_{1,t} | \mathbf{s}_{1,t}^{(n)})\bigg] - \bigg[-\sum_{p \in P_{a_{1,t}}^{+}} 
		p(a_{1,t} | \mathbf{s}_{1,t}^{(n)}, \mathbf{o}_{2,t}^{(n)}) 
		\log_2 p(a_{1,t} | \mathbf{s}_{1,t}^{(n)}, \mathbf{o}_{2,t}^{(n)})\bigg].
	\end{aligned}
\end{equation}
Our hypothesis is that manipulating TE will allow us to control the influence during social interactions. For instance, taking actions to increase TE could promote influence, which could sequentially enhance collaboration between agents. Ideally, being responsive and legible would aid all participants to achieve their goals during interactions. However, this strategy may become self-sacrificing or altruistic in some settings, a phenomena often explored in game theory \cite{james1993}. This happens when being legible conflicts with an agent's primary goal, or when an interaction is competitive.

Entropy measures the degree of uncertainty in actions: higher entropy means greater uncertainty, and lower entropy means less. Increasing $TE_{\mathbf{a}_{2,t}^{(n)}\rightarrow a_{1,t}}$ increases the uncertainty of actions when ignoring other agents and decreases the uncertainty when considering them (i.e., $H(P_{a_{1,t}}^{-})$ and $H(P_{a_{1,t}}^{+})$ in Equation \eqref{get_te} respectively), thus promoting influence. Conversely, decreasing TE resists this influence.

\subsection{Q-learning}

The TE calculations above require distributions over actions (a policy). We illustrate how these can be obtained using Q-learning \cite{watkins1989learning}. However, it is important to highlight that our approach is not restricted to Q-learning and could be used with other probabilistic policies obtained through reinforcement learning. We employ Temporal Difference learning \cite{sutton2018learning} to find Q-values representing the expected cumulative rewards for actions in specific states, following the Bellman equation \cite{bellman1957dynamic},
\begin{equation}
	\label{TD_update}
	\begin{aligned}
		Q(s,a) \leftarrow (1-\alpha)Q(s,a) + \alpha(r+\gamma \max_{a'}Q(s',a')).
	\end{aligned}
\end{equation}
Here $Q(s,a)$ is the Q-value for state $s$ and action $a$, $\alpha$ is the learning rate, $r$ is the immediate reward, $\gamma$ is the discount factor, $s'$ is the next state and $a'$ is the action in the next state. This iterative process enables the agent to learn the optimal Q-values over time, determining the most advantageous actions to take in various states. The optimal action $a^*$ at current state is determined by
\begin{equation}
	\label{opt_action}
	\begin{aligned}
		a^{*}=argmax_{a}Q(s, a).
	\end{aligned}
\end{equation}
In contrast to this deterministic policy, a probabilistic policy can be employed by converting the Q values to probabilities, then sampling from the distribution to choose the action. We do this by taking the softmax \cite{gibbs_elementary_2010} over Q-values at a given state.
\begin{equation}
	\label{softmax_n_sample}
	\begin{aligned}
		a^* \sim P_a = Softmax(Q(s,a)|_{s})
	\end{aligned}
\end{equation}
Q-learning often uses an epsilon-greedy strategy \cite{Sutton1998} to balance exploration and exploitation. This approach alternates between choosing actions with the highest Q-value and occasionally exploring other options with a small probability ($\epsilon$). We use this strategy in our implementations.

In the proposed multi-agent setting where states are not fully observable, $s$ in Q-learning (referring to $Q(s,a)$) becomes the states of ego-agent and observations of other agents' states in the scene. To be specific, $s$ comprises $\mathbf{s}_{1,t}^{(n)}$ and $\mathbf{o}_{2,t}^{(n)}$, which denote the states of $P1$ and observations of measurable state information from $P2$ at time $t, t-1, ..., t-n$ respectively. The action $a$ is simply the actions of the agent $P1$ at time $t$ (i.e., $a_{1,t}$), resulting in the expanded Q table,
\begin{equation}
	\label{Q-table}
	\begin{aligned}
		Q(s,a) = Q(a_{1,t}, \mathbf{s}_{1,t}^{(n)}, \mathbf{o}_{2,t}^{(n)}),
	\end{aligned}
\end{equation}
and probabilistic Q-learning policy 
\begin{equation}
	\label{Q-policy}
	\begin{aligned}
		a_{1,t}^{*} \sim P_{a_{1,t}} = Softmax(Q(a_{1,t} | \mathbf{s}_{1,t}^{(n)}, \mathbf{o}_{2,t}^{(n)})).
	\end{aligned}
\end{equation}
This distribution over actions allows us to compute TE using the Q table. First, for the entropy assuming influence, we can directly use the original Q table from Equation (\ref{Q-policy}) as the probability distribution over actions:
\begin{equation}
	\label{comm_Q_values}
	\begin{aligned}
		P_{a_{1,t}}^{+} = Softmax(Q(a_{1,t} | \mathbf{s}_{1,t}^{(n)}, \mathbf{o}_{2,t}^{(n)})).
	\end{aligned}
\end{equation}
In order to measure the entropy assuming no influence, the other agent's ($P2$'s) history in the Q table can be marginalised over: 
\begin{equation}
	\label{margin_Q}
	\begin{aligned}
		Q(a_{1,t},\mathbf{s}_{1,t}^{(n)})=\frac{1}{m}\sum_{\mathbf{o}_{2_t}^{(n)}}Q(a_{1,t}, \mathbf{s}_{1,t}^{(n)}, \mathbf{o}_{2,t}^{(n)}),
	\end{aligned}
\end{equation}
with $m$ the number of possible observations $\mathbf{o}_{2,t}^{(n)}$. Then the same process as above is repeated to compute the action distributions assuming no influence:
\begin{equation}
	\label{nocomm_entropy}
	\begin{aligned}
		P_{a_{1,t}}^{-} = Softmax(Q(a_{1,t} | \mathbf{s}_{1,t}^{(n)})).
	\end{aligned}
\end{equation}
We can now calculate the TE according to Equation~\eqref{get_te} and incorporate this into the reward term in the POMDP according to Equation~\eqref{add_te2reward}.

\subsection{Connections to Legibility}
The proposed TE augmentation is closely related to legibility. \citet{dragan_legibility_2013} formalise legible motion as motion that enables an observer to confidently infer the correct goal configuration after observing only a snippet of the trajectory. To do so, they propose a metric to evaluate legibility in a trajectory planning scenario, with the legibility L($\xi$) of a trajectory $\xi$ defined as: \[L(\xi)=\frac{\int{P(G^*|\xi)f(t)dt}}{\int{f(t)dt}},\] where $G^*$ is the actual goal across the trajectory. Trajectories are more legible if the probability $P(G^*|\xi)$ is higher. Consider an interactive scenario with two agents, agent 1 and agent 2. Denoting the trajectory or action of agent 1 by $\tau_1$ and agent 2 by $\tau_2$, and their goals $g_1$ and $g_2$, the legibility of agent 2 from the perspective of agent 1 can be expressed as: \[ L(g_2|\tau_2,\tau_1)=P(g_2|\tau_2,\tau_1). \] Using Bayes theorem, this probability can be expressed as: \[ P(g_2|\tau_2,\tau_1)=\frac{P(\tau_2|g_2,\tau_1)P(g_2|\tau_1)}{P(\tau_2|\tau_1)}. \] Now consider the TE from agent 1’s trajectory to agent 2’s trajectory, and for simplification, let us assume a history length of 1: 
\begin{equation}
	\begin{aligned}
		TE(\tau_1 \rightarrow \tau_2) &= H(\tau_2(t)|\tau_2(t-1)) - H(\tau_2(t)|\tau_2(t-1), \tau_1(t-1)) \\
		&= P(\tau_2(t), \tau_2(t-1), \tau_1(t-1)) \log \Bigg[ \frac{P(\tau_2(t)| \tau_2(t-1), \tau_1(t-1))}{P(\tau_2(t), \tau_2(t-1))} \Bigg].
	\end{aligned}
\end{equation}
This measures the amount of information that the past states of agent 1 ($\tau_1(t-1)$) provide about the future states of agent 2 ($\tau_2(t)$), over and above the information provided by the past states of agent 2 ($\tau_2(t-1)$) alone. The term \[ \log{\frac{P(\tau_2(t)| \tau_2(t-1), \tau_1(t-1))}{P(\tau_2(t), \tau_2(t-1))}} \] captures how much additional information about $\tau_2(t)$ is provided by $\tau_1(t-1)$ over and above $\tau_2(t-1)$. TE extends mutual information to consider the temporal dynamics and causality between the time series \cite{Schreiber2000}. Thus, TE also includes mutual information about how $\tau_1$ affects $\tau_2$. This affects both $P(\tau_2|g_2,\tau_1)$ and $P(\tau_2|\tau_1)$, which consequently affects the legibility $L(g_2|\tau_2,\tau_1)$. This relationship is particularly powerful, as TE gives us an implicit mechanism to modulate legibility, which is known to be challenging to measure directly.


\section{Experiment and Results}
\label{sec:experiment_n_results}

\begin{figure}[t]
	\centerline{\includegraphics[width=0.8\linewidth, trim=2.6cm 4.33cm 2.8cm 4.72cm, clip]{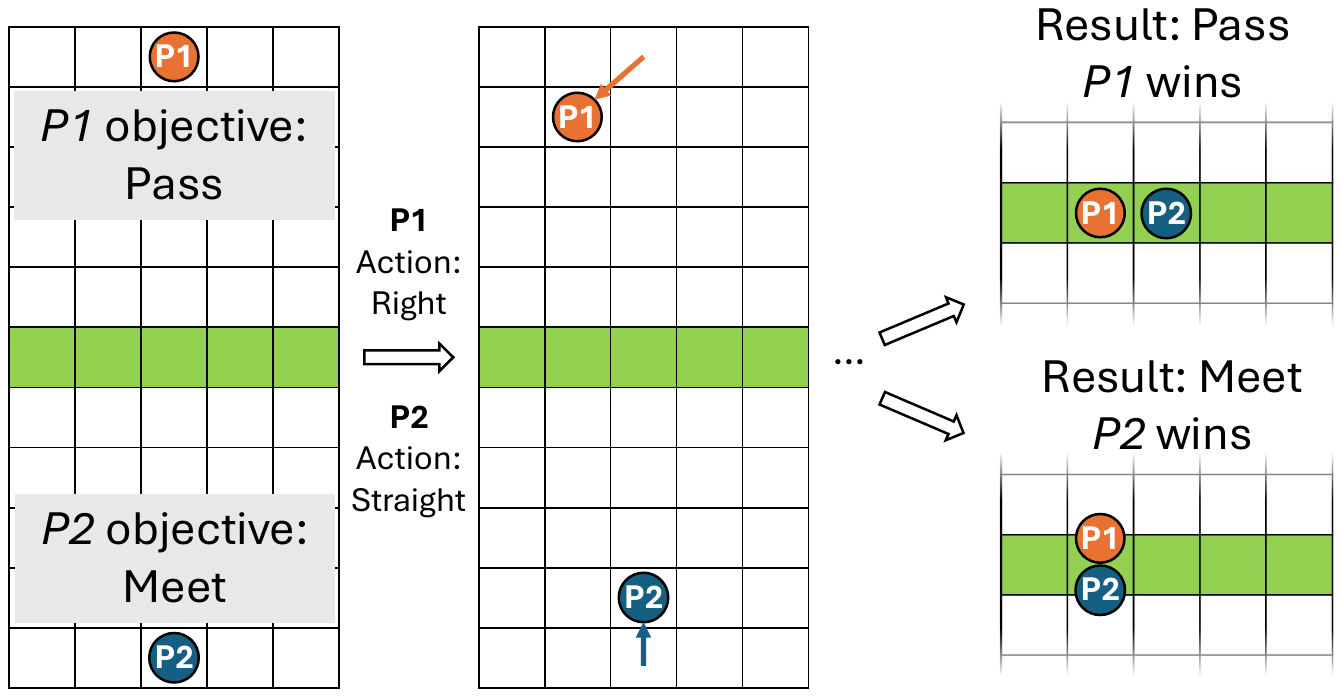}}
	\caption{A corridor dilemma episode: $P1$ is assigned to pass, and $P2$ is assigned to meet. In the first turn, $P1$ chose \textit{right}, and $P2$ chose \textit{straight}. After 5 turns, the result could be either \textit{pass} (as in the top scenario), making $P1$ the winner, or \textit{meet} (as in the bottom scenario), making $P2$ the winner.}
	\label{fig:corridor_demo}
\end{figure}

We investigate the proposed approach in our corridor dilemma setting (Fig. \ref{fig:corridor_demo}), a simplified 11x5 grid-world where two players start at opposite ends of a corridor at randomised starting positions with their own randomly assigned objectives: passing or meeting. In each round, players move forward and choose to move left, straight, or right, which means players cannot choose to stay at the same place or move backward. The game ends when both agents reach the same row, resulting in either a pass or a meeting, with scores based on predefined objectives. 

This game is a mixture of zero-sum (competitive) and collaborative games, requiring anticipation, strategy, and potential collaboration. We select this setting to mirror the broad range of scenarios a robot operating in the real world may encounter. These may be zero-sum (e.g., one agent wants to meet while the other does not) at times, or collaborative (e.g., both agents want to meet or pass) at others. The robot is never aware of the human's objective, and the human does not know the robot's objective.

We apply the proposed approach using agent coordinates as states and observations. In this specific application, the reward $r$ in Equation~\eqref{add_te2reward} is $r=0$ when the agent takes an action before reaching the middle row of the corridor, $r=10$ when the agent successfully achieves its objective, and $r=-10$ when it fails to achieve its objective after reaching the middle row. The TE reward components are normalised by min-max scaling according to the maximum possible TE before being linearly rescaled to between 0 and 10 to match the objective rewards. We note that TE is generally small so the game score still dominates the reward.

We conducted simulation, simulated human-agent experiments and real-world human-robot experiments to validate the proposed method. For all experiments, we set both the Q-learning rate and discount factor to $0.8$. These values were chosen empirically. To avoid memory constraints and improve computational speed, we use sparse matrices for the Q tables in Eq.~\eqref{margin_Q}. A history length of 5 is used. ($i.e., n=5$, the maximum history length.) Based on the TE reward scaling factor $\phi$, we define three types of agent: Non-TE agent ($\phi=0$, the baseline), Positive-TE agent ($\phi=10$, promoting influence) and Negative-TE ($\phi=-10$, resisting influence) agent.

\subsection{Simulation}
\label{subsec:simulation}

We conducted a range of simulations, putting each agent type against one another in a self-play framework where pairs of agents are simultaneously trained against each other for 30000 episodes with randomised objectives. 
$P1$ and $P2$ refer to the two players shown in Figure \ref{fig:corridor_demo}. For each experiment, we ran 30000 episodes and gathered results from six trials with distinct random seeds and computed their averages for presentation. All training runs were confirmed to converge within the specified number of steps. More simulation details can be found in Appendix \ref{apx:simulation_details}.

In addition to the self-play pairs, we established baseline tests for comparison. In each baseline test, a trained agent (Non-TE agent from the Non-TE vs.\,Non-TE pair, Pos-TE agent from the Non-TE vs.\,Pos-TE pair, Neg-TE agent from the Non-TE vs.\,Neg-TE pair) interacted with an agent controlled by a variant of the social force model initially proposed by Helbing and Molnar \cite{Helbing1995}. A wide variety of social force models have been proposed as extensions of the original formulation. These models can be broadly categorised as either non-anticipative, which do not explicitly model the intentions of surrounding agents~\cite{smith_ModellingContraflowCrowd_2009, farina_WalkingAheadHeaded_2017, jiang_ExtendedSocialForce_2017}, or anticipative, which incorporate such intention modelling~\cite{li_ImprovedSocialForce_2020, lu_PedestrianDynamicsMechanisms_2020, hu_AnticipationDynamicsPedestrians_2023}. Accordingly, we establish three simple and representative baselines that span the spectrum from zero intention knowledge to full intention knowledge. The following three models were implemented:
\begin{itemize}
	\item \textbf{Pure Social Force (Pure-SF)}: This is the simplest goal-driven model. When the agent’s objective is to meet the other agent, it selects the action that minimises the distance to the opponent. When the objective is to pass, it selects the action that maximises the distance. Pure-SF serves as a simple, intuitive policy reflecting straightforward reactive behaviour. The policy is defined as:
	\begin{equation}\label{eq:pureSF}
		a_t^*=
		\begin{cases}
			\operatorname*{arg\,min}\limits_{a_t \in A}(x_{t+1}^{self}-x_{t}^{op}), &\text{if Objective}=\text{meet} \\
			\operatorname*{arg\,max}\limits_{a_t \in A}(x_{t+1}^{self}-x_{t}^{op}), &\text{if Objective}=\text{pass}
		\end{cases},
	\end{equation}
	where $a_t$ denotes the current action chosen from the set of possible actions $A$, $x_{t+1}^{self}$ is the predicted lateral position of the social force agent, and $x_{t}^{op}$ is the current lateral position of its opponent.
	\item \textbf{Imperfect Knowledge Social Force (IPK-SF)}: This agent introduces uncertainty in opponent modelling. At each step, the agent has an 80\% chance of being informed of its opponent’s objective and assuming that the opponent follows the Pure-SF policy (Equation~\eqref{eq:pureSF}). In the remaining 20\% of cases, the agent makes a random guess of the opponent’s next position, $x_{t+1}^{op}$. IPK-SF simulates realistic scenarios in which an agent has incomplete or noisy information, reflecting limited situational awareness common in HRI. The resulting policy is:
	\begin{equation}\label{eq:IPKSF}
		a_t^*=
		\begin{cases}
			\operatorname*{arg\,min}\limits_{a_t \in A}(x_{t+1}^{self}-x_{t+1}^{op}), &\text{if Objective}=\text{meet} \\
			\operatorname*{arg\,max}\limits_{a_t \in A}(x_{t+1}^{self}-x_{t+1}^{op}), &\text{if Objective}=\text{pass}
		\end{cases},
	\end{equation}
	\item \textbf{Perfect Knowledge Social Force (PK-SF)}: This baseline assumes complete knowledge of the opponent’s objective and predicts its next position accordingly, essentially representing an upper-bound scenario for social force-based strategies. In this model, the agent always knows its opponent’s objective and assumes the opponent follows the Pure-SF policy (Equation~\eqref{eq:pureSF}), enabling it to predict the opponent’s next location. The policy formulation is identical to that of IPK-SF in Equation~\eqref{eq:IPKSF}.
\end{itemize}

Collectively, these baselines span a range of agent capabilities, from simple reactive (Pure-SF), to partially informed (IPK-SF), to fully informed (PK-SF), allowing us to compare our agents with the proposed framework to rule-based agents with varying levels of opponent knowledge. 

\begin{table*}[th!]
	\centering
	\caption{Final average success rate of competition (SRCP) and collaboration (SRCL) in simulation.}\label{tab:cc}
	\small
	\begin{tabular}{cccccc}
		\toprule
		\multirow{2}{*}{Experiment} & \multirow{2}{*}{Agent (TE type)} & \multicolumn{2}{c}{SRCP (\%)} & \multicolumn{2}{c}{SRCL (\%)}\\
		\cmidrule(lr){3-4} \cmidrule(lr){5-6}
		& & Mean & Std & Mean & Std\\
		\midrule
		\multirow{2}{*}{Random vs. Random}  & $P1$ (Random) & 50.00\% & - & \multirow{2}{*}{50.00\%} & \multirow{2}{*}{-}\\
		& $P2$ (Random) & 50.00\% & - \\
		\midrule
		\multirow{2}{*}{Non-TE vs. Non-TE}  & $P1$ (Non-TE) & 48.31\% & 7.42\% & \multirow{2}{*}{63.50\%} & \multirow{2}{*}{10.76\%}\\
		& $P2$ (Non-TE) & 51.69\% & 7.42\% \\
		\midrule
		\multirow{2}{*}{Non-TE vs. Pos-TE}  & $P1$ (Non-TE) & \textbf{63.38\%} & 7.62\% & \multirow{2}{*}{76.97\%} & \multirow{2}{*}{4.19\%}\\
		& $P2$ (Pos-TE) & 36.62\% & 7.62\% \\
		\midrule
		\multirow{2}{*}{Non-TE vs. Neg-TE}  & $P1$ (Non-TE) & 51.03\% & 8.16\% & \multirow{2}{*}{51.39\%} & \multirow{2}{*}{5.90\%}\\
		& $P2$ (Neg-TE) & 48.97\% & 8.16\% \\
		\midrule
		\multirow{2}{*}{Pos-TE vs. Pos-TE}  & $P1$ (Pos-TE) & 49.95\% & 6.47\% & \multirow{2}{*}{\textbf{91.72\%}} & \multirow{2}{*}{3.97\%}\\
		& $P2$ (Pos-TE) & 50.05\% & 6.47\% \\
		\midrule
		\multirow{2}{*}{Pos-TE vs. Neg-TE}  & $P1$ (Pos-TE) & 43.49\% & 7.96\% & \multirow{2}{*}{65.50\%} & \multirow{2}{*}{10.62\%}\\
		& $P2$ (Neg-TE) & 56.51\% & 7.96\% \\
		\midrule
		\multirow{2}{*}{Neg-TE vs. Neg-TE}  & $P1$ (Neg-TE) & 49.84\% & 9.35\% & \multirow{2}{*}{49.70\%} & \multirow{2}{*}{8.15\%}\\
		& $P2$ (Neg-TE) & 50.16\% & 9.35\% \\
		\bottomrule
	\end{tabular}
	\label{corridor_table_coll_n_comp}
\end{table*}

We report the averaged success rates for both competition and collaboration (denoted as SRCP - Success Rate for Competition and SRCL - Success Rate for Collaboration, respectively) to assess the agents' capacity to compete or collaborate effectively. In this context, competition refers to scenarios where agents are assigned different objectives, while collaboration refers to situations where both agents share the same objective. These success rates are quantified as the number of successful outcomes in either competition or collaboration, divided by the total number of instances of competition or collaboration. Specifically, the success rate is calculated as follows: 
\[
\text{Success Rate} = \frac{\text{Number of rounds that the player won}}{\text{Total number of rounds completed}}.
\]
Additionally, since each round could be either collaborative or competitive for the player, we compute the SRCL and SRCP by considering only those rounds where players were either collaborating or competing. Thus, the more rounds the players completed, the more stable the metrics became. The results are presented in Table \ref{corridor_table_coll_n_comp}. 

Across all experimental conditions, interactions involving at least one Positive-TE agent consistently achieve higher collaboration performance and fairer competitive outcomes than interactions without Positive-TE, indicating that Positive-TE agents positively influence their opponents’ behaviour.

In competition, the Non-TE agent in the Non-TE vs.\, Positive-TE experiment achieved the highest final performance with a 63.38\% SRCP, with a corresponding reduction in SRCP for the paired Positive-TE agent. This is as expected, since increasing influence or legibility can lead to the self-sacrificing or altruistic behaviour described in section~\ref{subsec:transfer_entropy}. This increase in influence allows the Non-TE agent to learn to satisfy its objectives more frequently both in circumstances of collaboration ($P1$ and $P2$ share the same objectives) and competition ($P1$ and $P2$ share different objectives).

Note that the Positive-TE vs.\,Positive-TE pair reaches the highest SRCL (91.72\%), and also much fairer outcomes in terms of competition (49.95\% vs. 50.05\%) compared to the Non-TE vs.\,Positive-TE pair (63.38\% vs. 36.62\%). In addition, the Negative-TE agent in Positive-TE vs.\,Negative-TE group that tries to resist influence does not achieve a higher SRCP than the Non-TE agent in the Non-TE vs.\,Positive-TE group. These findings suggest that engaging in either receiving or resisting influence may compromise performance, but training with a Positive-TE agent is beneficial for group collaboration and contributes to the fairness of the group. This is clearly evident as all pairs featuring at least one Positive-TE agent consistently attain better collaboration performance than those without, and results in fairer competition outcomes (SRCP) for the Positive-TE vs.\,Positive-TE pair.

\begin{table*}[th!]
	\centering
	\caption{Average success rate of competition (SRCP) and collaboration (SRCL) in  simulation against baselines.}\label{tab:cc_baseline}
	\small
	\begin{tabular}{cccccc}
		\toprule
		\multirow{2}{*}{Experiment} & \multirow{2}{*}{Agent (TE type)} & \multicolumn{2}{c}{SRCP (\%)} & \multicolumn{2}{c}{SRCL (\%)}\\
		\cmidrule(lr){3-4} \cmidrule(lr){5-6}
		& & Mean & Std & Mean & Std\\
		\midrule
		\multirow{2}{*}{Non-TE vs. Pure-SF}  & $P1$ (Non-TE) & \textbf{73.29\%} & 5.09\% & \multirow{2}{*}{43.09\%} & \multirow{2}{*}{6.90\%}\\
		& $P2$ (Pure-SF) & 26.71\% & 5.09\% \\
		\midrule
		\multirow{2}{*}{Non-TE vs. IPK-SF}  & $P1$ (Non-TE) & 67.68\% & 4.22\% & \multirow{2}{*}{44.52\%} & \multirow{2}{*}{3.69\%}\\
		& $P2$ (IPK-SF) & 32.32\% & 4.22\% \\
		\midrule
		\multirow{2}{*}{Non-TE vs. PK-SF}  & $P1$ (Non-TE) & 66.45\% & 4.77\% & \multirow{2}{*}{44.16\%} & \multirow{2}{*}{3.44\%}\\
		& $P2$ (PK-SF) & 33.55\% & 4.77\% \\
		\hline\hline
		\multirow{2}{*}{Pos-TE vs. Pure-SF}  & $P1$ (Pos-TE) & 47.06\% & 4.43\% & \multirow{2}{*}{73.63\%} & \multirow{2}{*}{9.20\%}\\
		& $P2$ (Pure-SF) & 52.94\% & 4.43\% \\
		\midrule
		\multirow{2}{*}{Pos-TE vs. IPK-SF}  & $P1$ (Pos-TE) & 35.73\% & 3.87\% & \multirow{2}{*}{\textbf{76.42\%}} & \multirow{2}{*}{3.81\%}\\
		& $P2$ (IPK-SF) & 64.27\% & 3.87\% \\
		\midrule
		\multirow{2}{*}{Pos-TE vs. PK-SF}  & $P1$ (Pos-TE) & 35.43\% & 4.25\% & \multirow{2}{*}{69.91\%} & \multirow{2}{*}{6.54\%}\\
		& $P2$ (PK-SF) & 64.57\% & 4.25\% \\
		\hline\hline
		\multirow{2}{*}{Neg-TE vs. Pure-SF}  & $P1$ (Neg-TE) & 39.55\% & 6.28\% & \multirow{2}{*}{69.82\%} & \multirow{2}{*}{9.00\%}\\
		& $P2$ (Pure-SF) & 60.45\% & 6.28\% \\
		\midrule
		\multirow{2}{*}{Neg-TE vs. IPK-SF}  & $P1$ (Neg-TE) & 36.56\% & 5.28\% & \multirow{2}{*}{69.13\%} & \multirow{2}{*}{7.67\%}\\
		& $P2$ (IPK-SF) & 63.44\% & 5.28\% \\
		\midrule
		\multirow{2}{*}{Neg-TE vs. PK-SF}  & $P1$ (Neg-TE) & 37.97\% & 8.74\% & \multirow{2}{*}{59.83\%} & \multirow{2}{*}{5.10\%}\\
		& $P2$ (PK-SF) & 62.03\% & 8.74\% \\
		\bottomrule
	\end{tabular}
	\label{corridor_table_coll_n_comp_baseline}
\end{table*}

The results of the baseline tests are presented in Table \ref{corridor_table_coll_n_comp_baseline}. Among the three baseline tests, the Pos-TE demonstrates consistently better collaborative performances against the baselines. The Pos-TE vs.\, IPK-SF pair achieves the highest SRCL (76.42\%), comparable to the performance of the None-TE vs.\, Positive-TE pair. However, the Positive-TE vs.\, Positive-TE pairs achieve substantially higher SRCL values. This indicates that the collaborative behaviour promoted by Positive-TE agents cannot be reproduced by simple social force models. In an ideal scenario, having two social force agents interact with each other could yield the best outcomes (i.e., 100\% SRCL and a 50\%-50\% SRCP), but such conditions are rarely achievable in practice. One has to assume that both agents share identical parameters, otherwise, the balance breaks. In practice, agents rarely have perfect knowledge of their partners, and social force models remain fundamentally passive, which behaves according to a fixed set of rules without the ability to actively adapt to new situations. By contrast, our proposed framework allows agents to proactively modulate information, enabling flexible adaptation to diverse and dynamic scenarios without prior knowledge or manually designed models.

These experiments clearly show that performance in interactive settings can be affected by controlling influence in basic social interactions, indicating potential applicability to broader human-robot-interaction scenarios. An ablation study can be found in Appendix \ref{apx:ablation}. Additional simulations, results and entropy analysis can be found in Appendix \ref{apx:additional_sim_res}, \ref{apx:mixed-TE} and \ref{apx:entropy_analysis}. Next, we verify the efficacy of the proposed method for virtual agents interacting with human users.

\subsection{Human-Agent Experiment}\label{subsec:human_agent_exp}

\begin{figure}[ht]
	\subfloat[Game interface design.]{\includegraphics[width=0.48\textwidth]{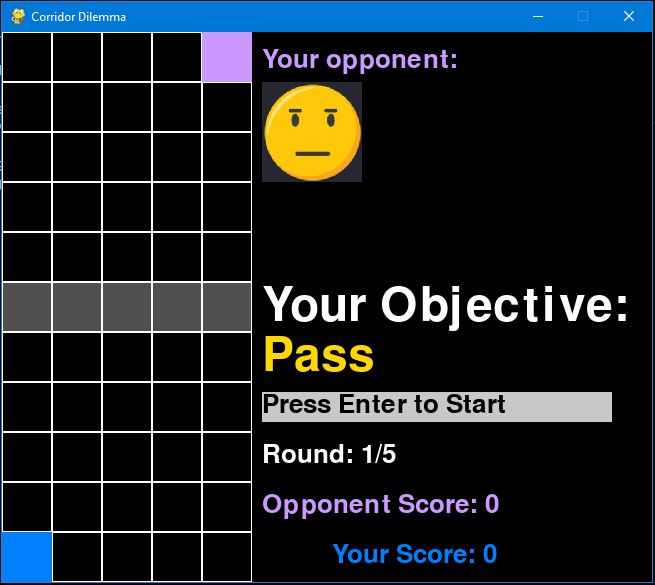}}%
	\hspace{0.03\textwidth}
	\subfloat[During playing. (Waiting for an action.)]{\includegraphics[width=0.48\textwidth]{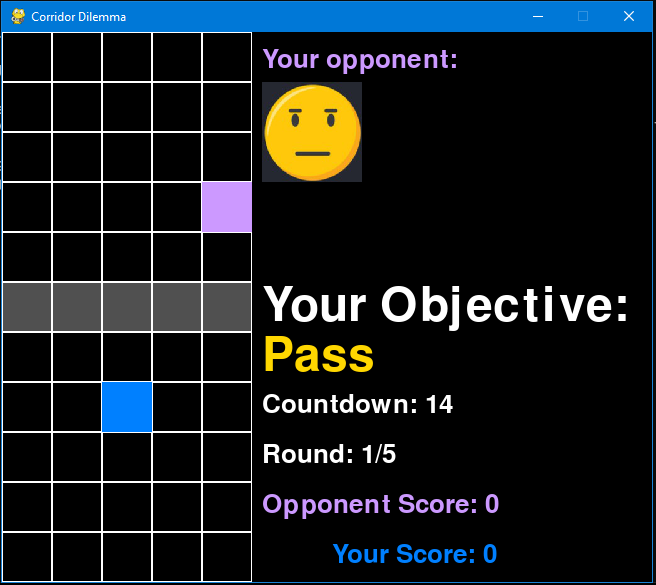}}%
	\caption{The interface design of \textit{the corridor dilemma} game.}\label{fig:game_interface_n_progress}
\end{figure}

\begin{figure}[ht]
	\subfloat[Success in round 1.]{\includegraphics[width=0.48\textwidth]{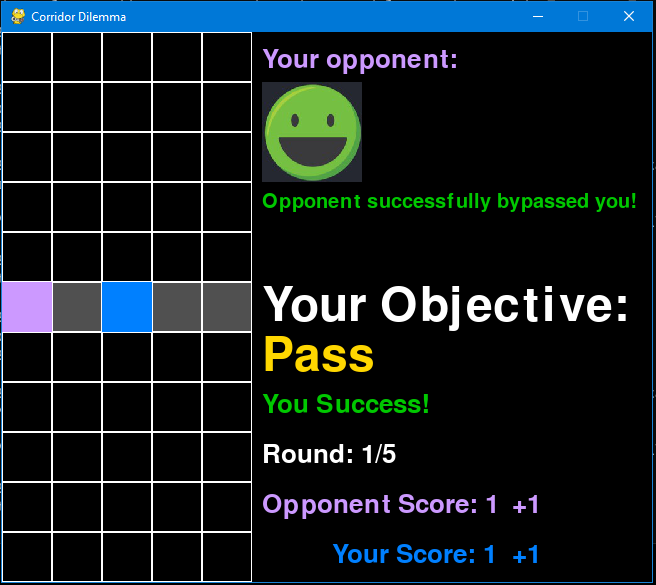}}%
	\hspace{0.03\textwidth}
	\subfloat[Fail in round 2.]{\includegraphics[width=0.48\textwidth]{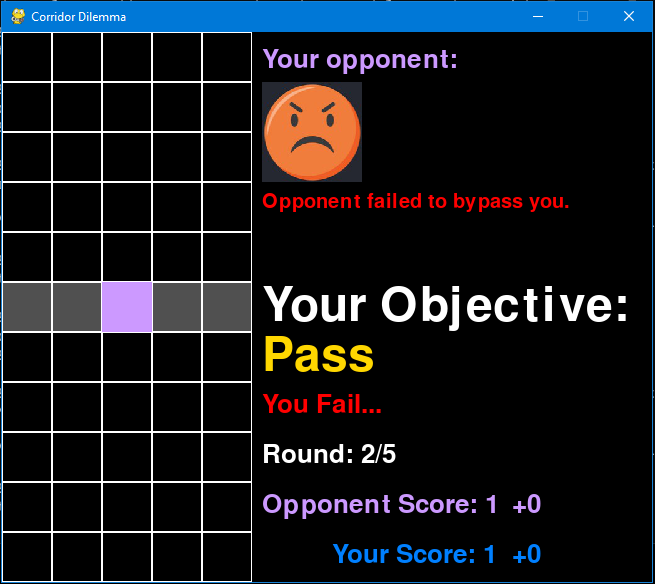}}%
	\caption{Example outcomes of the game. (Both players want to bypass each other.)}\label{fig:game_outcome}
\end{figure}

We conducted a user study\footnote{Approved by the Monash University ethics committee. Project ID: 41607} in a virtual corridor dilemma environment. The game interface design is displayed in Figures \ref{fig:game_interface_n_progress} and \ref{fig:game_outcome}. On the left side of the interface, an 11x5 grid corridor is shown, and participants can use the Up, Left, Right arrow keys to control their character, which is the blue tile that starts from a random location on the bottom row. The opponent, which is one of the trained RL agents introduced in \ref{subsec:simulation}, is the purple tile starting from a random location on the top row. The rules of game are the same as explained in \ref{sec:methodology}. On the right side of the interface, the opponent's response, the user's objective, the number of rounds and the scores are shown. From the start to the end of each trial, the participant needs to move five steps, and they have five seconds to think and enter an action for each step. If the five-second countdown is exceeded without user input, the user's character will automatically move straight for one step.

\begin{figure*}[ht]
	\centerline{\includegraphics[width=.8\linewidth]{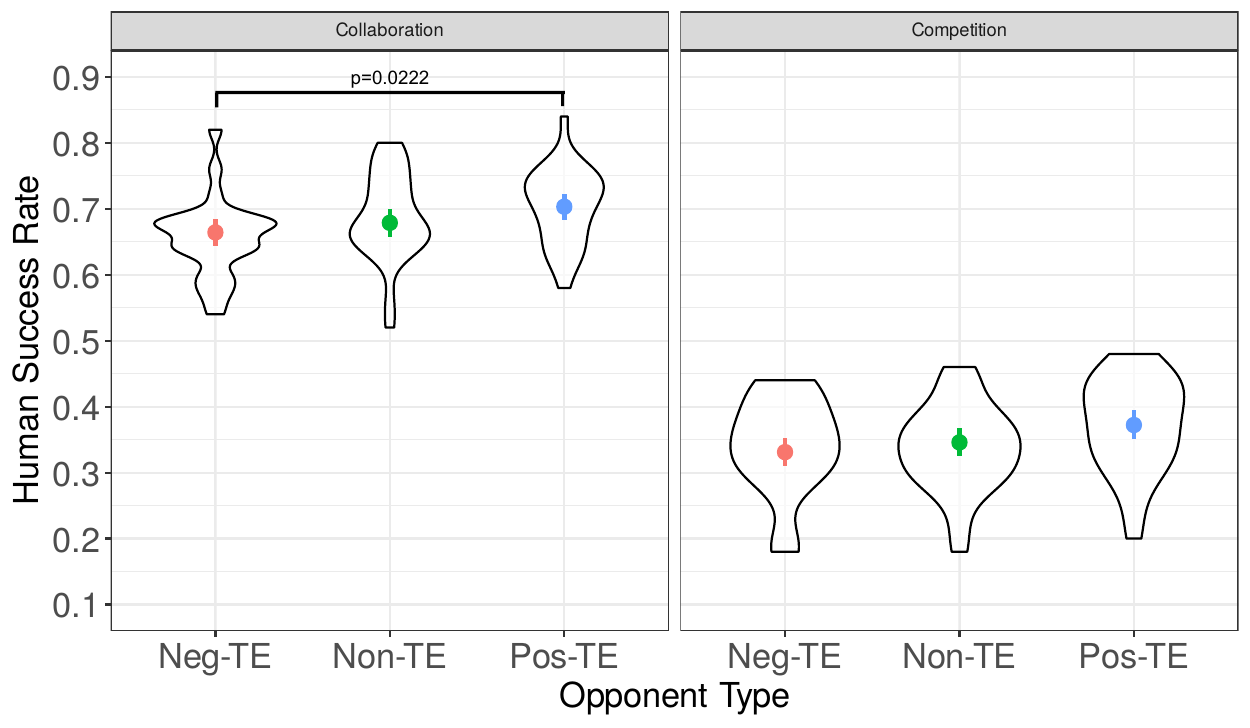}}
	\caption{Human-agent experiment results depicting collaboration and competition performance of humans in terms of their success rate. The dots and bars represent the means and 95\% confidence intervals respectively.}
	\label{fig:cc_us}
\end{figure*}

Here, human participants played against three agents. Given that humans may display ambivalence, selfishness or altruism, we assumed that a Non-TE agent, being ambivalent, would most accurately represent the typical human. Thus, participants are tested against three agents pre-trained under this assumption: 1. Non-TE agent ($P1$) vs. \textbf{Non-TE agent ($P2$)}, 2. Non-TE agent ($P1$) vs. \textbf{Positive-TE agent ($P2$)}, and 3. Non-TE agent ($P1$) vs. \textbf{Negative-TE agent ($P2$)}. Each participant was asked to play against each of the three pre-trained agents (randomly ordered) for 100 consecutive rounds with randomly generated objectives. The number of rounds was selected to provide ample opportunity for each participant to familiarise themselves with an agent and allow user performance to converge to a stable level. As part of the study, participants were told that one of the opponents could potentially be a real human player. A power analysis indicated that a minimum of 24 participants was required to achieve 80\% statistical power. We recruited 26 participants (11 female, 15 male, aged between 20 and 37). Their collaboration and competition performances were observed across 100 rounds of gameplay against each opponent, and subsequently averaged across all participants. These findings are illustrated in Figure~\ref{fig:cc_us}.

\begin{table}[ht]
	\centering
	\caption{Pairwise contrasts of opponent type within each collaboration condition from the GLMM of the Human-Agent experiment. Estimates are on the log-odds scale. Odds ratios (OR) are shown for interpretability. P-values are Tukey-adjusted for multiple comparisons.}
	\label{tab:human-agent-glmm}
	\begin{tabular}{cccccc}
		\toprule
		Condition                      & Contrast        & Estimate (log-odds) & z     & p (Tukey) & Odds Ratio (OR)        \\ \midrule
		\multirow{3}{*}{Collaboration} & Neg-TE - Non-TE        & -0.14               & -1.71 & 0.20      & 0.87            \\
		& Neg-TE - Pos-TE        & -0.22               & -2.65 & \textbf{0.022}     & \textbf{0.80}            \\
		& Non-TE - Pos-TE        & -0.08               & -0.94 & 0.62      & 0.92            \\ \midrule
		\multirow{3}{*}{Competition}   & Neg-TE - Non-TE        & 0.01                & 0.13  & 0.99      & 1.01            \\ 
		& Neg-TE - Pos-TE        & -0.14               & -1.68 & 0.21      & 0.87            \\
		& Non-TE - Pos-TE        & -0.15               & -1.80 & 0.17      & 0.86            \\ \bottomrule
	\end{tabular}
\end{table}

\begin{figure*}[t]
	\centerline{\includegraphics[width=.8\linewidth]{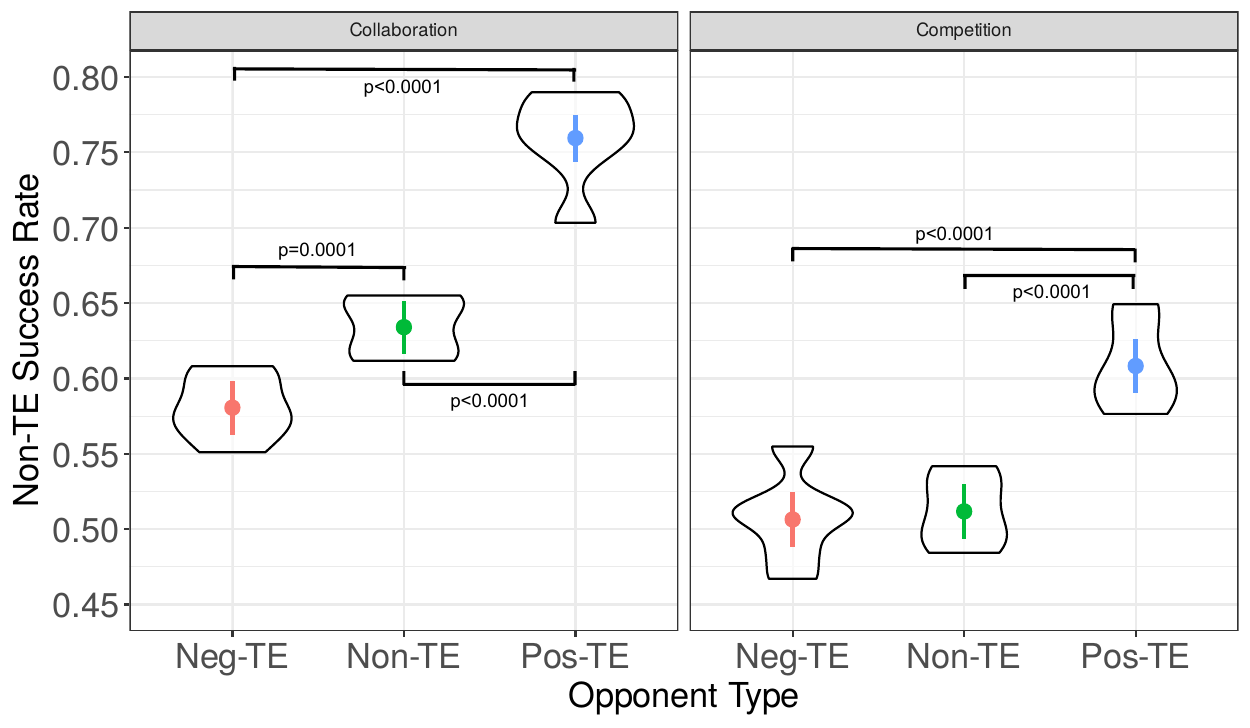}}
	\caption{Simulation results of Non-TE agents against three agent types in terms of Non-TE agents' success rate aligning with the experimental results with humans (Figure \ref{fig:cc_us}). The dots and bars represent the means and 95\% confidence intervals respectively.}
	\label{fig:cc}
\end{figure*}

\begin{figure}[ht]
	\centering
	\includegraphics[width=0.8\linewidth, trim=0.5cm 0.0cm 0.5cm 0.0cm, clip]{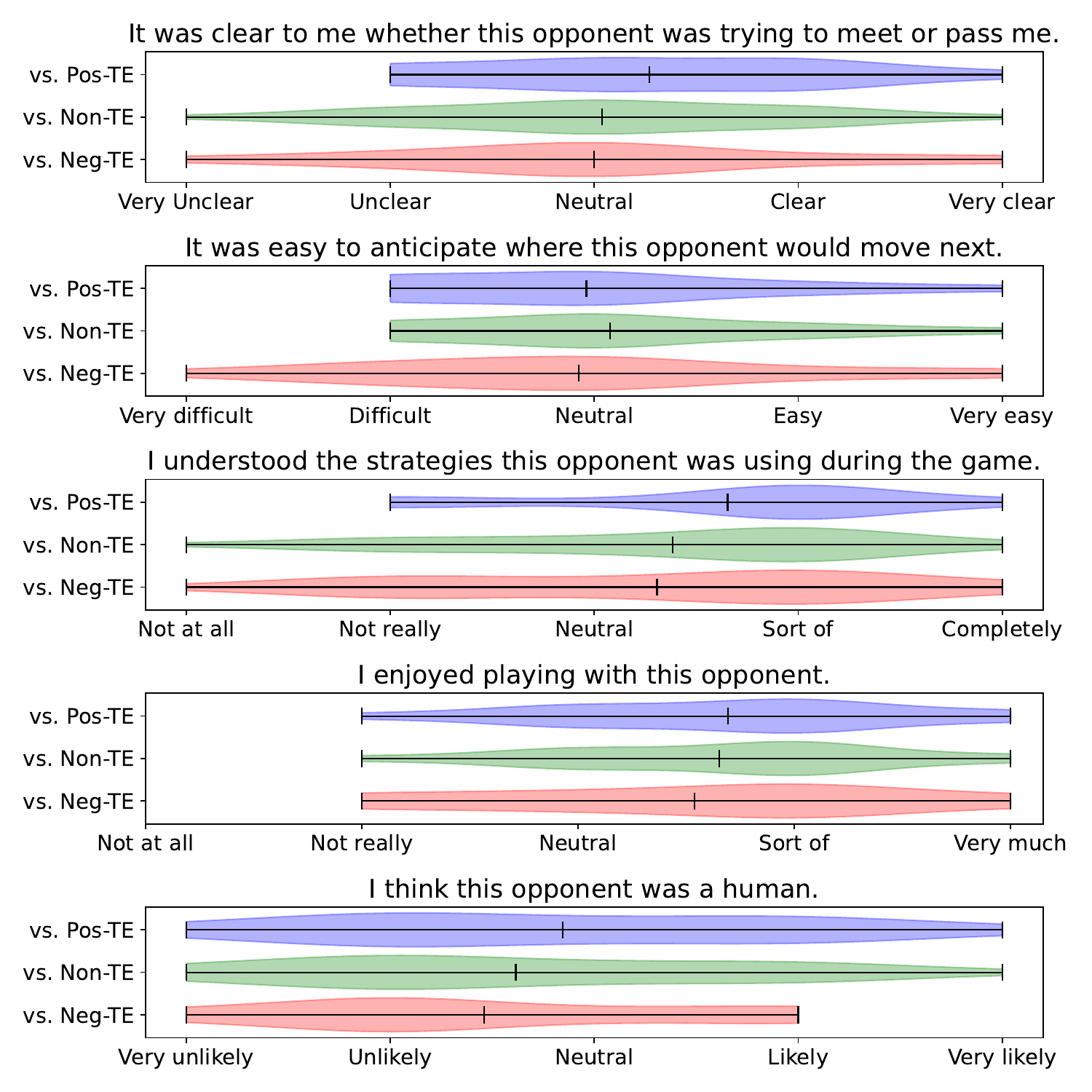}
	\caption{Full survey results of the human-agent experiments, where mean values are marked.}
	\label{fig:survey}
\end{figure}

To handle random effects such as within-individual variability, we fitted a generalised linear mixed model (GLMM) with a binomial family and logit link to examine the effect of opponent type on the probability of a success outcome for humans, including random intercepts for participant. Results showed a significant difference among the three opponent types ($\chi^2 = 9.624,\, p = 0.008$), whereas no significant effect was observed for trial order ($\chi^2 = 0.356,\, p = 0.551$), indicating the absence of ordering effects such as fatigue. Pairwise contrasts of opponent type within each condition from the GLMM is shown in Table \ref{tab:human-agent-glmm}. Estimates are on the log-odds scale. Odds ratios (OR) are shown for interpretability. P-values are Tukey-adjusted for multiple comparisons. Under collaborative conditions, participants were significantly more likely to succeed against Positive-TE opponents compared with Negative-TE opponents ($p = 0.022$), corresponding to an OR of 0.80. No other pairwise contrasts were significant (all $p > 0.17$).

However, Figure \ref{fig:cc_us} shows a clear visual trend, supported by the odds ratios in Table \ref{tab:human-agent-glmm}, indicating that participants were more likely to succeed when playing against a Positive-TE opponent compared with the other opponent types in both collaboration and competition. This seems to be highly desirable behaviour in robots, agents should support humans to achieve their goals in collaboration, and behave transparently (and hence altruistically) when in competition. A Negative-TE agent resists influence, leading to decreased collaboration performance among human players. This results in competitive performance that remains relatively stable when compared to playing against Non-TE agents. Importantly, these real world findings match the observed trend from the simulations conducted in Section~\ref{subsec:simulation}. Figure~\ref{fig:cc} shows the performances of a trained Non-TE agents when facing the three different types of agents used in the human study. (The pairwise contrasts table for simulation can be found in the Appendix \ref{apx:pairwise_contrasts_in_sim}.) It should be noted that the simulation settings and Q-learning agents are considerably less complex than real human behaviour. Consequently, the effect of the TE reward in these simulations is notably more pronounced than in the user study, though both follow the same trends. Overall, these findings suggest that manipulating influence through TE does indeed impact human behaviour. By boosting or resisting influence, we can design agents that facilitate improved or diminished performance in human participants.

After each interaction, participants were asked to give feedback via a short survey about their experience with the agent. Figure \ref{fig:survey} presents survey findings indicating participants perceive the Pos-TE agent as more legible and human-like. One-way ANOVA tests \cite{girden1992anova} showed no significant differences suggesting only subtle perceptual distinctions. Combined with the significantly different performance results in Figure \ref{fig:cc_us}, this implies that our method enables implicit influence modulation, or implicit communication. Humans are only moderately aware of differences between policies, despite clear quantitative performance differences.

\subsection{Human-Robot Experiment}
\label{subsec:human_robot_experiment}

\begin{figure}[ht]
	\centerline{\includegraphics[width=0.5\linewidth]{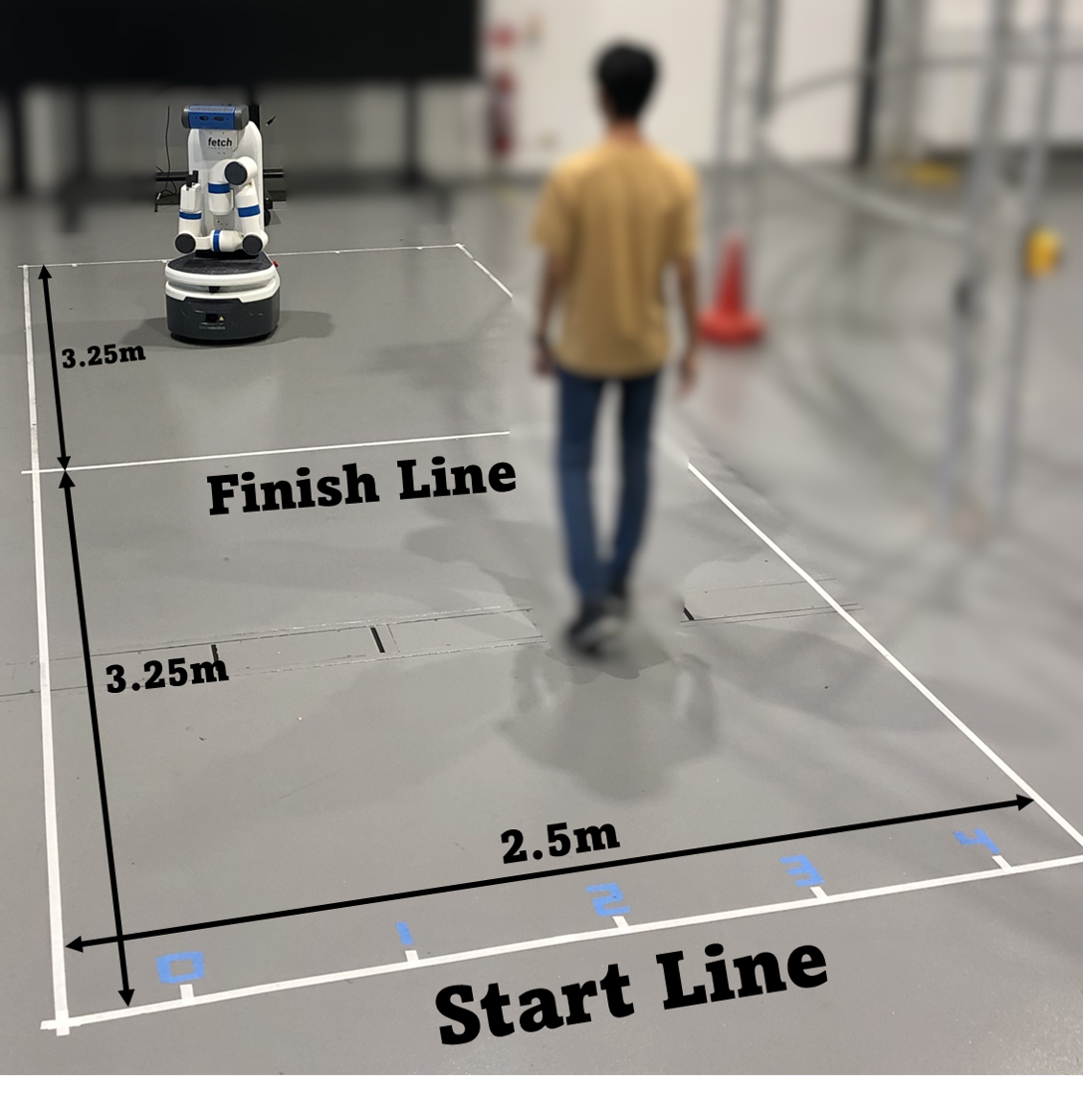}}
	\caption{Human-robot experiment setting. The region measures 2.5 meters in width and extends 6.5 meters in length. Starting positions are labelled as 0 to 4 on the start line.}
	\label{fig:human-robot-setting}
\end{figure}

The results above show that implicit communication via TE can influence human behaviour in virtual worlds, allowing for greater collaboration and affecting performance in competition, but it is unclear to what extent this holds with a physical robot. To see if the real robot morphology would alter previous results, the corridor dilemma was translated into a physical environment\footnote{Approved by the Monash University ethics committee. Project ID: 45746} (Figure \ref{fig:human-robot-setting}). We applied the pre-trained policies to a Fetch robot \cite{Wise2016FetchF}, which was chosen due to its semi-humanoid appearance and flexible manoeuvrability. The Time Elastic Bands (TEB) planner \cite{roesmann2012} was selected to ensure smooth navigation trajectories. We utilised the on-board lidar to scan for human legs, enabling real-time tracking of the human's position and mapping it to the corresponding grid world coordinates. By inputting both the human and robot's grid world locations into a pre-trained Q-table, the robot could plan its actions (forward, left, or right) in real time. Given the discrete nature of the Q-table, the robot executes the next action only after moving one grid unit longitudinally. Although the algorithm uses discrete representations, motion primitives render the discrete actions continuous during the experiment, so that from the participant’s perspective, the robot navigates smoothly in a continuous manner. In the human-agent experiment, both participants moved synchronously, but in practice, this is difficult to achieve. To address this, during action planning, the robot calculates the ideal human position along the longitudinal axis and substitutes the actual human longitudinal position with this ideal value, ensuring that the robot and human remain symmetrically aligned along the finish line. As a result, the robot continues moving smoothly without pausing for its human opponent. We instructed participants to maintain a pace similar to that of the robot.

The experiment settings remained consistent with those of the human-agent virtual experiment (Section \ref{subsec:human_agent_exp}), with the exception that the grid was not explicitly marked on the floor, and continuous motion used instead of  turn-based moves. Participants were instructed to navigate the corridor on foot in a smooth and continuous manner. For each trial, participants began at a randomly assigned starting position and pursued a randomly allocated objective: either to "meet" (stop in front of the robot) or to "pass" (avoid "meet"). Participants were required to walk slowly and steadily towards the finish line, avoiding large strides or lateral movements, while striving to achieve their designated objective. Although the robot navigated using a grid-world-based algorithm, its trajectory was smoothed to generate continuous natural motion. 

\begin{figure}[ht]
	\centerline{\includegraphics[width=0.7\linewidth]{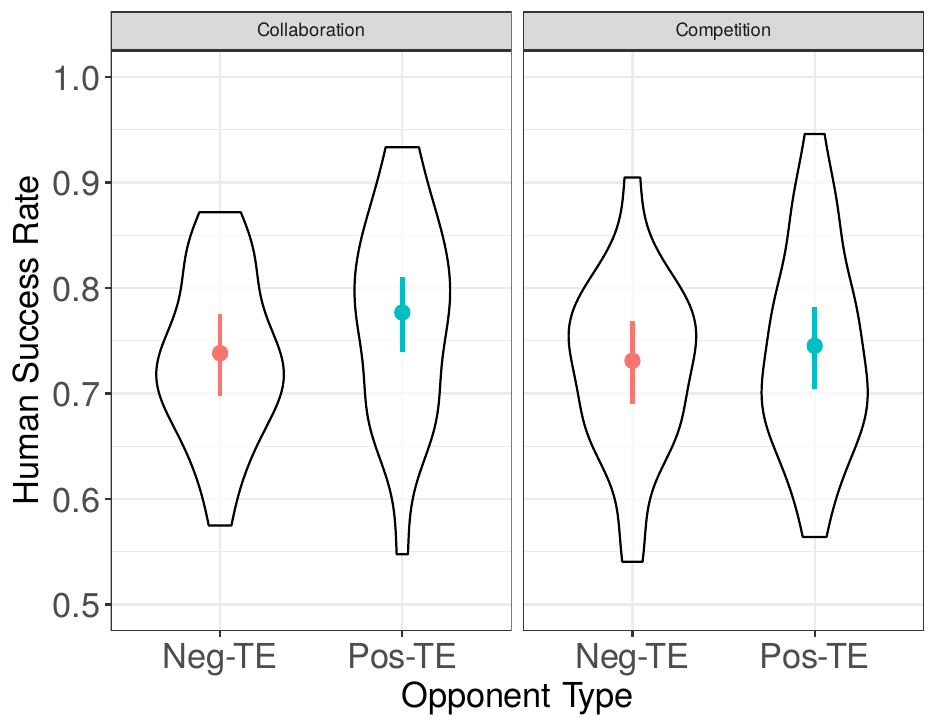}}
	\caption{Human-robot experiment results depicting collaboration and competition performance of humans in terms of their success rate. The dots and bars represent the means and 95\% confidence intervals respectively.}
	\label{fig:cc_human_robot_exp}
\end{figure}

Participants interacted with two agents pre-trained under the assumption that a typical human is represented by a Non-TE agent: 1. Non-TE agent ($P1$) vs. \textbf{Positive-TE agent ($P2$)}, and 2. Non-TE agent ($P1$) vs. \textbf{Negative-TE agent ($P2$)}. Each participant was asked to play against the two agents (randomly ordered) for 80 consecutive rounds with randomly generated objectives. The number is selected to minimise physical strain on the participants. We recruited 21 participants (7 women, 14 men, aged between 24 and 34) and asked them to walk towards or pass the robot, following randomly generated objectives. Their collaboration and competition performances were observed across 80 rounds of walking using each policy (policies tested successively, order randomised), and subsequently averaged across all participants. Similar to the human–agent experiment, we fitted a GLMM to analyse the results. Results showed no significant difference between the two opponent types ($\chi^2 = 3.293,\, p = 0.07$), and no significant effect was observed for trial order ($\chi^2 = 1.827,\, p = 0.176$), indicating the absence of ordering effects such as fatigue. The findings are illustrated in Figure~\ref{fig:cc_human_robot_exp} and Table \ref{tab:human-robot-glmm}.

\begin{table}[ht]
	\centering
	\caption{Pairwise contrasts of opponent type within each collaboration condition from the GLMM of the Human-Robot experiment. Estimates are on the log-odds scale. Odds ratios (OR) are shown for interpretability. P-values are Tukey-adjusted for multiple comparisons.}
	\label{tab:human-robot-glmm}
	\begin{tabular}{cccccc}
		\toprule
		Condition & Contrast & Estimate (log-odds) & z & p (Tukey) & Odds Ratio (OR) \\ \midrule
		Collaboration & Neg-TE - Pos-TE & -0.21 & -1.90 & 0.057 & 0.81 \\
		Competition   & Neg-TE - Pos-TE & -0.07 & -0.66 & 0.51 & 0.93 \\ \bottomrule
	\end{tabular}
\end{table}

In general human performance against the physical robot was much better than the results obtained against the virtual agent. Figure \ref{fig:cc_human_robot_exp} and Table \ref{tab:human-robot-glmm} show that participants achieve a near significantly higher SRCL ($p = 0.057$) when interacting with a Pos-TE robot compared to a Neg-TE robot. This effect aligns with our earlier finding that Pos-TE trained robots are desirable for human-robot collaboration. 

However, the results also reveal that participants achieve a slightly higher SRCP when interacting with a Neg-TE robot (OR = 0.93), deviating from findings in previous sections. Several factors could account for this discrepancy. First, we note that the shift away from a more strategic turn-based setting to a continuous motion setting reduces the competitive nature of the game, and introduces a number of potentially confounding variables. Humans are advantaged in the physical world experiment as they are able to move faster, and there are potentially many more variables and factors providing motion cues than the grid-based representation that robot policies used. Moreover, factors such as the physical presence of the robot, and proxemics effects overriding the assigned human objective are also likely to play a role. For example, participants may believe that they have achieved their goal of meeting a robot, while their personal space preferences may lead our system to label the interaction otherwise. These factors may result in outliers in the relatively small sample sized dataset, in particular for human participants that do not follow our assumption of the average human behaving like a non-TE agent. This is also potentially exacerbated by fatigue, as participants were required to perform 160 trials in an approximately 90 minute session. 

\begin{figure}[t]
	\centering
	\includegraphics[width=0.85\linewidth, trim=0.4cm 0.0cm 0.45cm 0.0cm, clip]{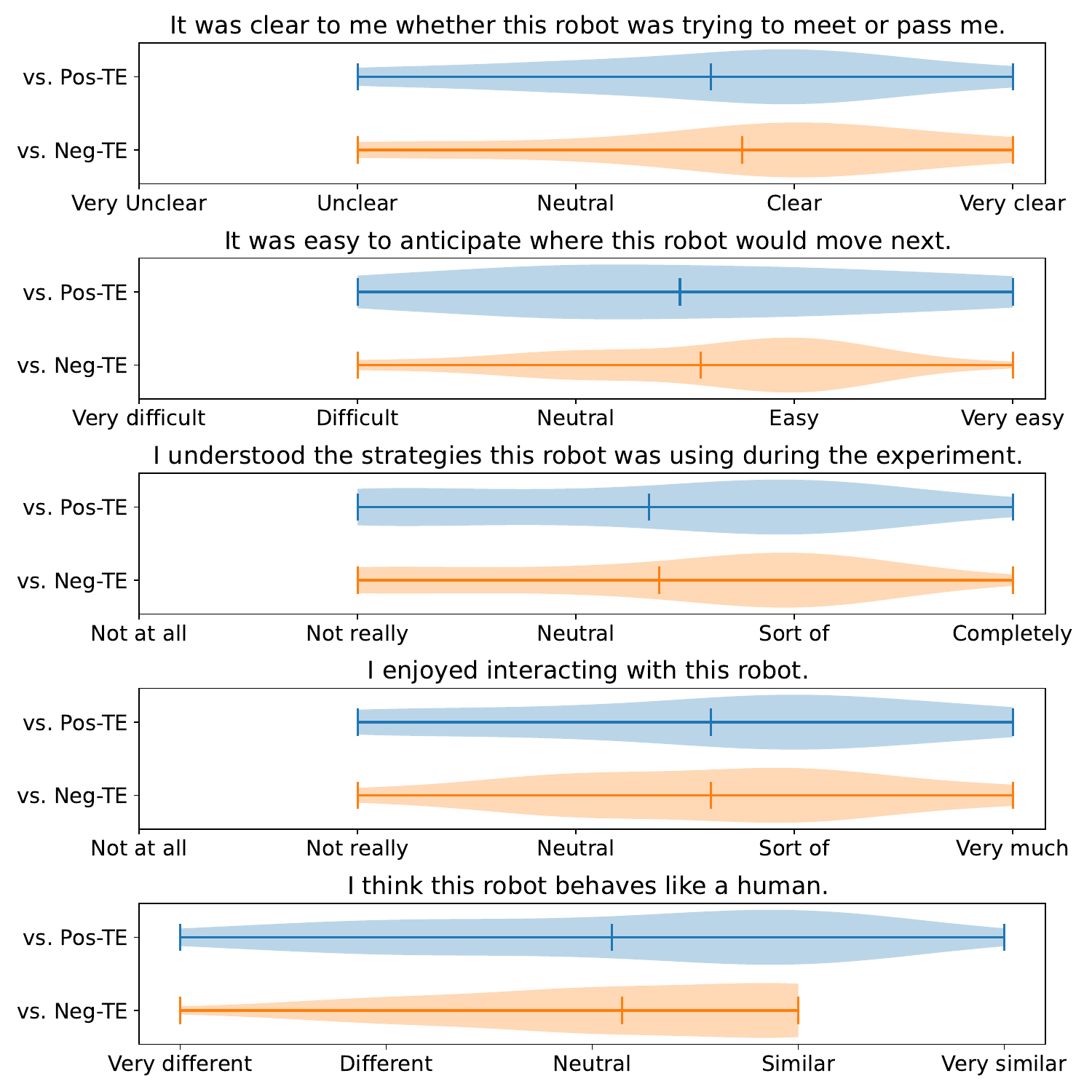}
	\caption{Partial survey results of the human-robot experiment, where mean values are marked by short vertical bars.}
	\label{fig:survey_human-robot}
\end{figure}

Despite these potential confounding factors, it is clear that TE reward augmentation can influence human behaviour and collaboration/ competition in real world interactions with robots. We also evaluated this qualitatively, with participants asked to complete a set of survey questions similar to those in the virtual human-agent experiment. Figure \ref{fig:survey_human-robot} presents survey questions. Statistical tests did not reveal significant differences, indicating only subtle perceptual differences with robot morphology.

\section{Extension in DRL and Multi-Agent Settings}

We have demonstrated the performance of the proposed framework in a setting with a discretised state space and interactions limited to two agents in Section \ref{sec:experiment_n_results}. To generalise the framework to more complex tasks, we discuss its extension to DRL and multi-agent settings and demonstrate its functionality in the Highway environment from the \textit{highway-env} collection \cite{highway-env}, a more complex environment featuring a continuous state space and multi-agent interactions.

\subsection{Handling Continuous Space}\label{subsec:handle_cs}

In continuous space, two primary challenges arise: calculating the action entropy and marginalising the continuous policy. For the first challenge, the general approach is to obtain the action distribution. In discrete action spaces, as in standard Deep Q-Learning (DQL) \cite{mnih2015human}, discrete Proximal Policy Optimisation (PPO) \cite{schulman2017proximal} and Soft Actor-Critic (SAC) \cite{haarnoja2018soft}, this can be achieved by applying a softmax over the Q-values for all possible actions. In continuous action spaces, as in continuous PPO and SAC, the policy network typically outputs a typically Gaussian distribution, \( a \sim \pi_{\theta}(\cdot|s) \), that can be directly used to compute the Shannon’s differential entropy assuming that the distributions are Gaussian:

\begin{equation}
	\label{eq:diff_entropy}
	H(\mathbf{x}) =-\int p(\mathbf{x})\log p(\mathbf{x}) \,d\mathbf{x}
	=\frac{D}{2}(1+\log (2\pi)) + \frac{1}{2}\log |\mathbf{\sigma}|.
\end{equation}

For the second challenge, there are a few possible approaches to approximate the marginalised policy. The most direct is to train a separate policy network using only partial observations. However, this is computationally expensive, particularly for deep networks. An alternative is to train a single policy network with input masking, but it remains costly in terms of training time and demands careful hyperparameter tuning and mask design. 

A third option is Monte Carlo estimation, which is the method we have adopted. This technique also requires only a single policy network. By sampling the partial observations designated as the source signal, feeding these samples into the policy network, and averaging the resulting action distributions, one can approximate the marginalised policy distribution. This approach is straightforward to implement, generally less computationally demanding than the other two approaches, and capable of producing reliable approximations. As the number of samples increases, the estimated distribution should, in theory, converge to the true marginalised distribution. This follows from the law of large numbers, which guarantees that the sample mean of independent and identically distributed (i.i.d.) random variables converges to the expected value. In this case, for a given full observation \(o\), let the source component be \(s\) and the remaining partial observation be \(o_{\backslash s}\). By sampling \(s \sim p(s)\), feeding \(o_{\backslash s} \cup s\) into the policy \(\pi(a|o)\), and averaging over \(N\) samples, we estimate:
\begin{equation}
	\label{eq:Monte-Carlo-estimation}
	\hat{\pi}(a|o_{\backslash s})=\frac{1}{N}\sum_{i=1}^{N}\pi(a|o_{\backslash s}, s^{(i)}).
\end{equation}
As \(N\rightarrow \infty\), this estimate converges to the true marginal:
\begin{equation}
	\label{eq:estimate_converge}
	\pi(a|o_{\backslash s})=\mathbb{E}_{s\sim p(s)} [ \pi(a|o_{\backslash s}, s^{(i)}) ],
\end{equation}
assuming \(\pi(a|o)\) is measurable and the samples are drawn i.i.d.\! from the correct distribution.

\subsection{Experiment and Results}

\begin{figure}[ht]
	\centerline{\includegraphics[width=0.7\linewidth]{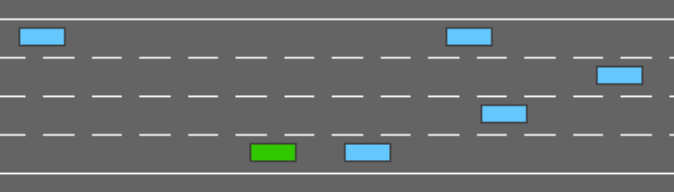}}
	\caption{The default Highway task. The green rectangle represents the ego-vehicle, and the blue rectangles represent other vehicles.}
	\label{fig:highway}
\end{figure}

The \textit{highway-env} \cite{highway-env} is a collection of environments for decision-making in Autonomous Driving. In this experiment, we demonstrate the proposed framework with a DQL setting using the Highway environment from this collection. In the Highway task, the ego-vehicle is driving on a multilane highway populated with other vehicles. Figure \ref{fig:highway} shows a single frame of this environment.

At each timestep, the ego vehicle observes its own location and velocity, as well as the relative locations and velocities of the surrounding vehicles. It can then choose one of five possible actions: Idle, Accelerate, Decelerate, Lane Left, or Lane Right. The other vehicles are controlled by the same Intelligent Driver Model (IDM) \cite{IDM2000} for longitudinal behaviours and the same Minimising Overall Braking Induced by Lane change (MOBIL) model \cite{MOBIL2007} for lateral behaviours.

The DQL approach with Monte Carlo approximation proposed in Section \ref{subsec:handle_cs} is used to train the ego-agent with the objective to match the speed of the other vehicles while avoiding collisions, while driving on the right side of the road is also rewarded. The reward function is expressed as:
\begin{equation}\label{eq:highway-reward}
	R(s,a)=a \left( 1-\frac{|v_{target} - v_{ego}|}{\Delta v_{max}} \right) +b\,\text{collision} + c \left( 1-\frac{|\text{Lane}_{ego}-\text{Lane}_{right}|}{N_{lane}}\right) + d\,r_{TE},
\end{equation}
where $v_{ego}$ denotes the speed of the ego-vehicle, $v_{target}$ denotes the common target speed of the other vehicles, $\Delta v_{max}$ denotes the maximum possible speed difference between $v_{ego}$ and $v_{target}$. Term $\text{collision}$ is $1$ if collision happens, and $0$ otherwise. $\text{Lane}_{ego}$ and $\text{Lane}_{right}$ are the index of the lane where the ego vehicle is currently driving on and the index of the right most lane respectively. $N_{lane}$ is the total number of lanes, which is three in our case. $r_{TE}$ is the normalised TE reward term, which is, borrowing denotions in Eq. \eqref{eq:estimate_converge}, computed as:
\begin{equation}
	r_{TE}=\frac{H(\pi(a|o_{\backslash s})) - H(\pi(a|o_{\backslash s}, s)}{\log|A|},
\end{equation}
where $o_{\backslash s}$ represents the partial observations of just the state of the ego-vehicle, and $s$ represents the observations of the other vehicles. $\log|A|$ is the theoretical maximum TE with $|A|$ being the number of possible actions. The $a,b,c,d$ in Eq. \eqref{eq:highway-reward} are coefficients for each reward term. We chose $a=0.4,\,b=-1,\,c=0.1$ in this experiment. By varying the coefficient of the TE reward, $d$, we aim to examine the effect of the proposed framework.

\begin{figure}[ht]
	\centerline{\includegraphics[width=\linewidth]{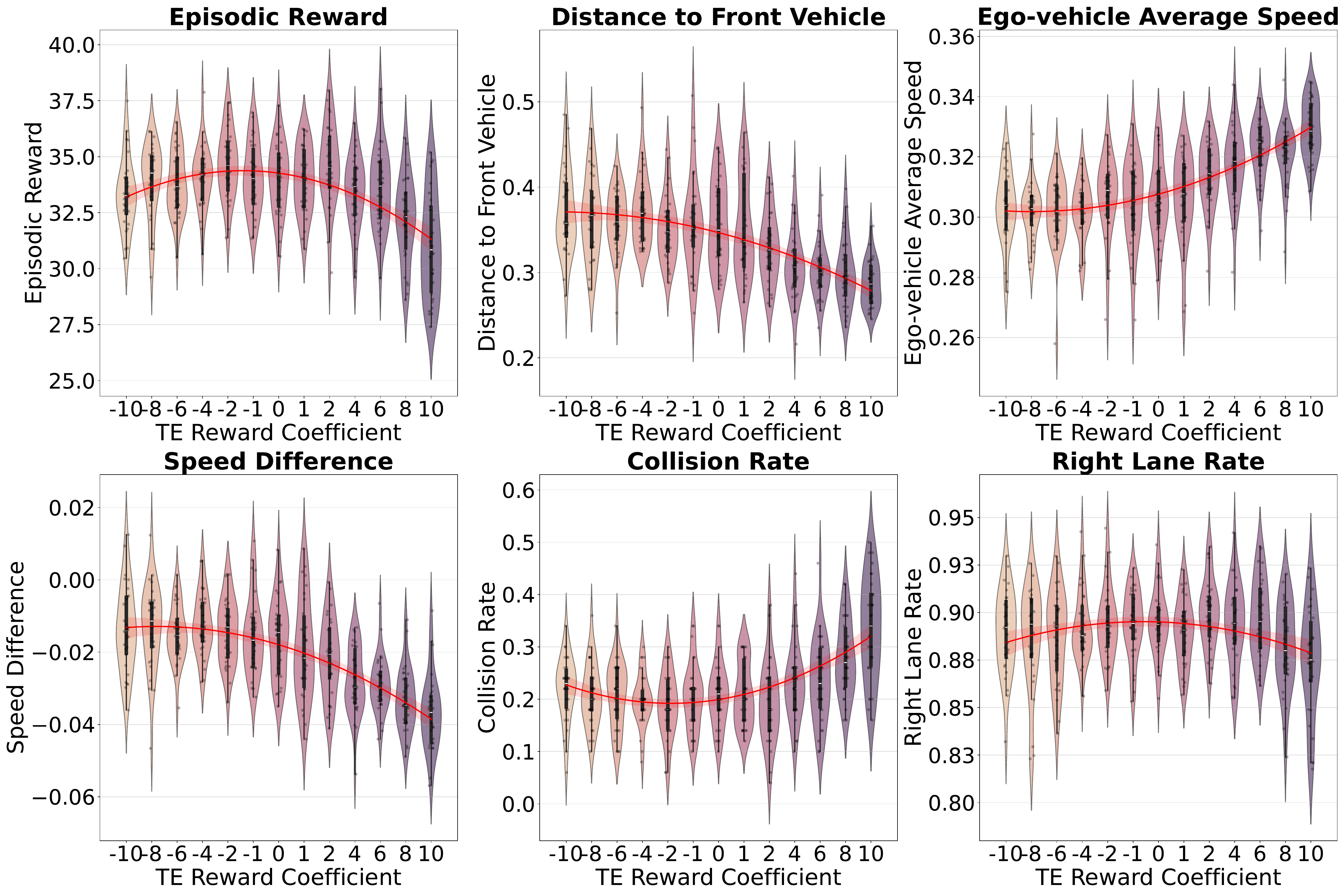}}
	\caption{Evaluation results of the TE reward coefficient sweeping across the range $[-10,10]$ with quadratic regressions (red curves) to show the trend. From left to right and top to bottom: episodic reward (excluding TE rewards), distance to the vehicle in front of the ego vehicle, average ego-vehicle speed, the speed difference (i.e., $v_{target} - v_{ego}$), the collision rate and the right lane rate.}
	\label{fig:highway-eval}
\end{figure}

The value of $d$ was swept across the range $[-10,10]$ in increments of two. This range was chosen empirically to ensure that the TE term did not dominate the overall objective reward, knowing that the raw TE value is inherently small. The model was trained using 30 random seeds for each value of $d$ and tested accordingly, yielding 30 data points per metric for each $d$ value. For analysis, we evaluated the following metrics during testing: episodic reward (excluding TE rewards), distance to the vehicle in front of the ego vehicle, average ego-vehicle (forward) speed, the (forward) speed difference (i.e., $v_{target} - v_{ego}$), the collision rate and the right lane rate, which specifies how often the ego vehicle stays on the right most lane. The evaluation results are presented in Figure \ref{fig:highway-eval}.

The episodic rewards show little variation when $d$ is roughly between -2 to 2. For values of $d$ outside this range, the overall reward decreases, with positive $d$ reducing the reward faster than negative $d$. The other plots indicate that positive-TE vehicles tend to maintain a shorter gap to the front vehicles and drive at higher speeds compared to negative-TE and non-TE vehicles. Consequently, they are less able to match the speeds of other road vehicles, resulting in a higher collision rate. These behaviours are consistent with the positive-TE reward, which encourages influence and promotes interaction and social engagement. For instance, the ego vehicle may accelerate to approach other vehicles (controlled by the MOBIL model \cite{MOBIL2007}) from behind to trigger lane changes. We provide an additional analysis in Appendix \ref{apx:te-vs-dist} to support this claim, illustrating how TE varies with inter-vehicle distance. A lower right-lane usage rate compared to non-TE vehicles also suggests that positive-TE vehicles are more interactive.

In contrast, negative-TE vehicles, which suppress influence, behave more conservatively, maintaining lower speeds and larger front gaps. However, at lower $d$ values ($d<-4$), the ego vehicle may make more irrational decisions, as indicated by the increasing collision rate and decreasing right-lane usage. While negative-TE vehicles prefer social independence, excessive suppression can cause the agent to make decisions that largely ignore the states of other vehicles.

These results indicate that the proposed influence modulation framework is effective in DRL settings. Positive-TE vehicles tend to be more interactive and assertive, promoting influence, while negative-TE vehicles behave more conservatively, suppressing influence. Unlike the corridor dilemma, where collaboration and competition are clearly separated, the Highway task involves a mixture of both dynamics. At first glance, it may appear collaborative, as all vehicles aim to drive at speed and avoid collisions, but each agent ultimately acts in its own interest. This highlights how the effect of influence modulation depends on the scenario. Appropriate use of influence can lead to beneficial behaviours, such as improved communication or altruism in the corridor dilemma with positive-TE. In contrast, inappropriate use may cause issues, exemplified by the more radical driving behaviour observed in the Highway task with positive-TE.

\section{Discussion}
The results in corridor-dilemma simulations show that the collaborative behaviour enabled by Positive-TE agents cannot be reproduced by rule-based social force models, which remain passive even under strong assumptions about partner knowledge. In contrast, our framework enables proactive information modulation, allowing agents to adapt without prior knowledge or hand-crafted models.

The experiments above validate that the proposed framework can influence human behaviour and increase or decrease human performance. Choosing appropriate TE augmentations to create a situation of information asymmetry for a human-robot interaction is an interesting philosophical decision, depending on the level of agency we want our robots to achieve. In economics and game theory, information asymmetry is a powerful theory explaining the effects of transactions where incomplete information can tip the balance of power in favour of another agent. The experiments above suggest that we can exploit this, potentially using TE to bias our robots towards an Asimov-like reframing \cite{anderson2008asimov}:

\begin{itemize}
	\item A robot should cede to a human when their objectives conflict.
	\item A robot should seek to fulfil its own objectives, as long as these do not conflict with the law above.
\end{itemize}

Although we demonstrate our framework in social navigation and autonomous driving scenarios, we believe it is a general approach applicable to social influence modulation across diverse human-robot interaction tasks. The framework is adaptable to various interactions, for example human-robot handover, autonomous driving and natural language interactions, due to it's behaviour-model-free nature. 

The effect of influence is context-dependent. In some scenarios, such as the corridor dilemma, promoting influence can lead to positive outcomes, however, excessive influence may be detrimental, as observed in the Highway task. This highlights that the modulation of influence must be carefully tailored to the specific environment and objectives. Choosing the appropriate level of influence according to context is therefore essential for achieving desirable agent behaviour. For instance, in tasks such as the corridor dilemma, where increasing interactions between agents to gain information is desirable, incorporating positive-TE is a suitable choice. In contrast, in tasks such as the Highway environment, where increased interactions are generally discouraged due to potential risks, incorporating negative-TE may encourage the agent to behave in a safe and conservative manner. Importantly, the TE objective allows us to build beneficial behaviours into the robot model while also considering other design choices regarding the impacts of information asymmetry in a given setting.

\section{Limitations and future Work}
\label{sec:limitations_n_future}

Our influence modulation framework, demonstrated in a two-agent setting and a multi-vehicle autonomous driving setting, has shown potential for fostering complex cooperative and competitive scenarios and expressing different characteristics. A key benefit of the proposed framework is that it is behaviour-model-free because, as shown in Eq. \ref{action_distribution}, it only requires the ego-agent’s history states $\mathbf{s}_{1,t}^{(n)}$ and its observations of the other agents $\mathbf{o}_{2,t}^{(n)}$ and no explicit model of behaviour. In our case, this is a trajectory of state observations or the basic kinematics of vehicles. This allows the approach to learn directly from human and robot state data in real-world settings, with minimal assumptions. A simple reward augmentation can then enhance or suppress 
influences in human-robot interactions without explicitly identifying influencing factors or relying on explicit human behaviour models. However, this limits the interpretability of the robot behaviour, and potentially raises questions about what type of TE reward augmentation should be used for a desired response, which may be dependent on the level of altruism, selfishness or ambivalence displayed by a human collaborator. Future work investigating whether the proposed framework can be extended to infer the type of agent a human most closely resembles (neg-TE, non-TE, pos-TE) may help with appropriate reward augmentation selection.

As is the case with all human-robot experiments, there may be a number of confounding factors influencing our experimental results, from sample size to fatigue. This work lays the groundwork for expanding transfer entropy reward augmentation, but there is still room for significant exploration of these factors in alternative applications and extended settings. In addition, as suggested by our human-robot corridor dilemma experiments, deeper investigations regarding how robot morphology affects influence modulation will be an interesting avenue of future work.

The differing effects of the proposed framework in the corridor dilemma and the Highway task indicate that not all interactions are beneficial in every situation. Future work could explore methods to classify interaction scenarios to guide the application of TE rewards. Another limitation highlighted by these findings is that the current framework can only generally promote or suppress influence or interaction as a whole, whereas in some cases, targeting specific aspects of the interaction may be preferable. This could potentially be achieved if TE were computed with respect to specific reward signals or value functions, which remains a direction for future research.

While the inclusion of explicit communication could further enrich the interaction dynamics, our current focus is on implicit exchanges that arise naturally without explicit signalling. Exploring how explicit and implicit communication can be combined within the same framework would be an interesting direction for future research. In addition, another interesting direction for future research is to integrate more explicit Theory of Mind (ToM) mechanisms into the framework, allowing agents to reason about others’ beliefs and goals more directly.


\section{Conclusion}
\label{sec:conclusion}
This work proposed a framework to promote implicit communication for social HRI via influence modulation, which does not require explicit modelling or pre-existing knowledge of human participants. We test the proposed framework in simulations, human-agent and human-robot corridor dilemma experiments. Results indicate that a positive or negative TE reward can enhance or reduce participant performance based on whether the interaction is collaborative or competitive. We further demonstrate the applicability of the proposed framework in a DRL setting using the Highway task. The results show that the TE reward can either promote or suppress interactions, depending on its sign. Notably, a positive TE reward fosters collaboration and interaction in appropriate application scenarios, which is highly desirable for HRI. Overall, the framework requires only observations of measurable state information from other participants, and limited domain knowledge or assumptions to produce human-centric and interactive behaviours. As a result, we believe that this approach has significant potential for broader robotic applications.



\bibliography{references}

\appendices

\section{Simulation details}
\label{apx:simulation_details}
The detailed settings during self-play simulation are summarised as follows. We set both the learning rate and discount factor to 0.8. (i.e., $\alpha=\gamma=0.8$ in Equation \eqref{TD_update}.) The history length is set to 5. (i.e., $n=5$ in Equation \eqref{Q-table}-\eqref{nocomm_entropy}.) We adapted the $\epsilon$-decay strategy during training, where $\epsilon$ is calculated as:

\begin{equation}
	\label{eps-decay}
	\begin{aligned}
		\epsilon=\max(-\frac{1}{\textit{max. iteration}}\times \textit{current iteration} + 1, 0),
	\end{aligned}
\end{equation}

where the $\textit{max. iteration}=30000$ as mentioned in Section \ref{subsec:simulation}. This encourages agents to explore more at the beginning, then slowly converge to the optimal strategy during training.


\section{Ablation studies}
\label{apx:ablation}

\begin{figure}[ht]
    \centering
    \subfloat[Agent performances for the Non-TE (P1) vs. Pos-TE (P2) pair when sweeping the TE scaling factor.]
    {
        \includegraphics[width=0.48\linewidth]{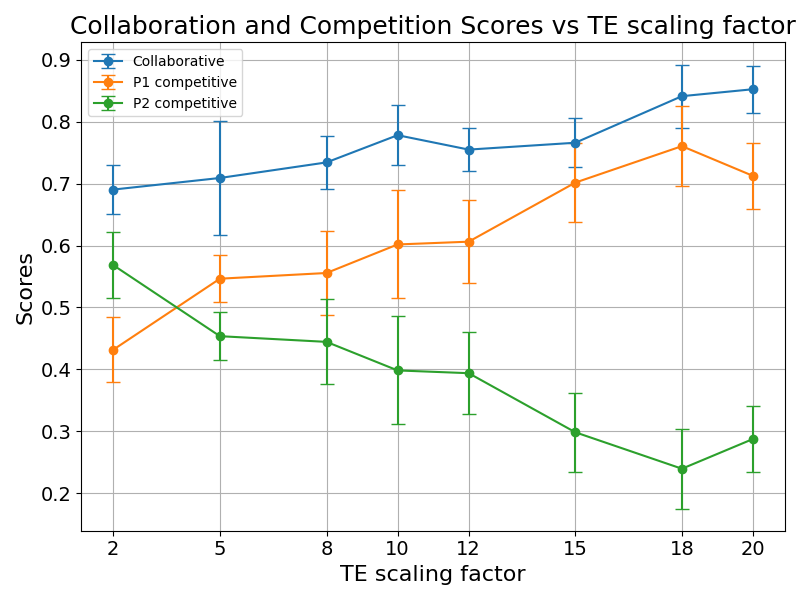}
        \label{fig:phi_sweep}
    }
    \hfill
    \subfloat[Agent performances for the Non-TE (P1) vs. Pos-TE (P2) pair when sweeping the history length.]
    {
        \includegraphics[width=0.48\linewidth]{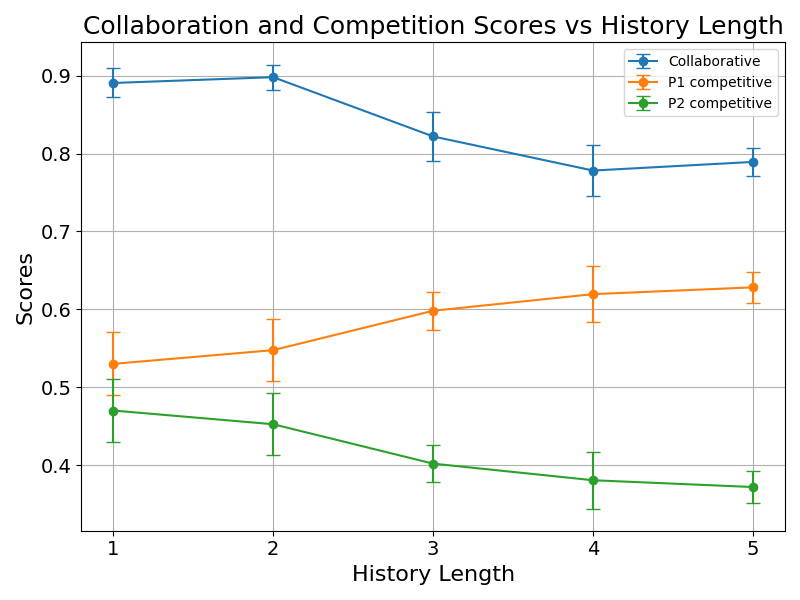}
        \label{fig:hist_len_sweep}
    }
    \caption{Comparison of agent performances under different parameter sweeps.}
    \label{fig:combined_sweeps}
\end{figure}

We conducted ablation studies to investigate the effects of the TE scaling factor and history length in the corridor dilemma experiment. We focused on the Non-TE vs. Pos-TE pair, sweeping the TE scaling factor from 2 to 20 and the history length from 1 to 5, while keeping the other parameter constant. The results are shown in Figures \ref{fig:phi_sweep} and \ref{fig:hist_len_sweep}. Higher TE scaling factors appear to enhance collaboration and promote stronger altruistic behaviour in the Pos-TE agent. In our main experiment, we selected a factor of 10 as a moderate value to avoid domination by the TE reward. Shorter history lengths lead to higher collaboration scores (i.e., SRCL) but less distinctive altruistic behaviour in the Pos-TE agent. We chose a history length of 5, as a longer history allows the agent to utilise the full information from past interactions. In practice, selecting these parameters may require preliminary studies to identify the most appropriate combination.

\section{Additional simulation results}
\label{apx:additional_sim_res}

In addition to measuring the success rate of collaboration and competition, we have also measured other performance metrics. The additional averaged performance for each agent has been summarised in Table \ref{tab:additional}. In order to facilitate a fair comparison between different pairs of agents, we proposed a metric called the \textit{Collective Performance Score} (CPS) which measures not only the individual performances of agents but also their effectiveness as a group. Denoting the success rate of passing and meeting as \textit{SRP} and \textit{SRM} respectively, and the baseline success rate of passing and meeting as \textit{BSRP} and \textit{BSRM} respectively, CPS is measured as follows:

\begin{equation}
	\label{CPS}
	\begin{aligned}
		CPS = &\frac{1}{2}((1-BSRP_{P1})SRP_{P1} + (1-BSRM_{P1})SRM_{P1}) \\
		+&\frac{1}{2}((1-BSRP_{P2})SRP_{P2} + (1-BSRM_{P2})SRM_{P2})
	\end{aligned}
\end{equation}

In scenarios where both agents take random actions at each turn, the probability of meeting each other is $BSRM=\frac{1}{5}*\frac{1}{5}*5=0.2$. The probability of passing each other is $BSRP=1-0.2=0.8$. These probabilities serve as a baseline. The highest SRP (94.43\%) and SRM (49.13\%) are both achieved by the Pos-TE agent $P1$ in the Pos-TE vs. Pos-TE pair, which also achieves the highest CPS (0.57). When comparing their SRP and SRM to the Non-TE vs. Pos-TE pair, although the Non-TE agent $P1$ has achieved similar results (93.86\% and 45.66\%), the performance of its paired Pos-TE agent is not as strong. This indicates that when both agents promote influence, a fairer game can be achieved. In addition, we observed that if an ego-agent promotes influence for itself as a recipient, it can also become a stronger influencer. This is evidenced by the higher TE measurements of agents when paired with a Pos-TE agent compared to when paired with a Non-TE or Neg-TE agent. These findings suggest that promoting influence not only benefits individual agents but also contributes to the fairness of the group. Promoting influence, therefore, is advantageous for everyone involved, fostering a more cooperative and equitable environment.

\begin{table*}[!htp]\centering
	\caption{Additional simulation results.}\label{tab:additional}
	\scriptsize
	\begin{tabular}{cccccccccc}\toprule
		\multirow{2}{*}{Experiment} &\multirow{2}{*}{Agent (TE type)} &\multicolumn{2}{c}{SRP (\%)} &\multicolumn{2}{c}{SRM (\%)} &\multicolumn{2}{c}{Averaged TE (bits)} &\multirow{2}{*}{CPS} \\\cmidrule{3-8}
		& &Mean &Std &Mean &Std &Mean &Std & \\\midrule
		\multirow{2}{*}{Random vs. Random} &P1 (Random) &80.00\% &- &20.00\% &- &- &- &\multirow{2}{*}{0.32} \\
		&P2 (Random) &80.00\% &- &20.00\% &- &- &- & \\
		\midrule
		\multirow{2}{*}{Non-TE vs. Non-TE} &P1 (Non-TE) &86.70\% &9.39\% &26.91\% &11.70\% &0.56 &0.36 &\multirow{2}{*}{0.39} \\
		&P2 (Non-TE) &86.57\% &9.41\% &26.58\% &12.13\% &0.55 &0.36 & \\
		\midrule
		\multirow{2}{*}{Non-TE vs. Pos-TE} &P1 (Non-TE) &93.86\% &7.13\% &45.66\% &13.59\% &0.75 &0.36 &\multirow{2}{*}{0.49} \\
		&P2 (Pos-TE) &80.96\% &10.94\% &32.98\% &12.59\% &1.3 &0.29 & \\
		\midrule
		\multirow{2}{*}{Non-TE vs. Neg-TE} &P1 (Non-TE) &84.50\% &9.99\% &21.56\% &11.48\% &0.44 &0.37 &\multirow{2}{*}{0.34} \\
		&P2 (Neg-TE) &84.21\% &10.42\% &21.33\% &12.03\% &0.64 &0.29 & \\
		\midrule
		\multirow{2}{*}{Pos-TE vs. Pos-TE} &P1 (Pos-TE) &\textbf{94.43}\% &6.46\% &\textbf{49.13}\% &15.45\% &1.44 &0.18 &\multirow{2}{*}{\textbf{0.57}} \\
		&P2 (Pos-TE) &91.15\% &8.49\% &46.73\% &14.57\% &1.43 &0.18 & \\
		\midrule
		\multirow{2}{*}{Pos-TE vs. Neg-TE} &P1 (Pos-TE) &83.78\% &10.45\% &25.32\% &12.34\% &1.25 &0.33 &\multirow{2}{*}{0.39} \\
		&P2 (Neg-TE) &88.31\% &9.95\% &29.72\% &13.39\% &0.8 &0.24 & \\
		\midrule
		\multirow{2}{*}{Neg-TE vs. Neg-TE} &P1 (Neg-TE) &80.42\% &11.44\% &21.40\% &12.42\% &0.47 &0.23 &\multirow{2}{*}{0.34} \\
		&P2 (Neg-TE) &81.69\% &11.36\% &22.06\% &11.67\% &0.46 &0.21 & \\
		\bottomrule
	\end{tabular}
\end{table*}


\section{Mixed-TE Agents}
\label{apx:mixed-TE}

In addition to the self-play simulation experiments discussed in Section \ref{subsec:simulation}, we trained agents with mixed TE types. Specifically, we assigned one agent in the corridor dilemma a random scaling factor $\phi$ for the TE reward, randomly selecting $\phi$ from 0, 10, and -10 for each episode. This agent is termed a Mixed-TE agent. The other agent retained a fixed TE scaling factor. We trained Non-TE, Pos-TE, and Neg-TE agents against Mixed-TE agents using the same settings described in Section \ref{apx:simulation_details}. The results for agents trained against the Mixed-TE agent presented in Table \ref{tab:mixed_result}, are not as good as those for agents trained against the Non-TE agent presented in Table \ref{tab:additional}. 

Training with a mixed agent results in an overall decrease in both SRM (Success Rate of Meeting) and SRP (Success Rate of Passing) for all agents. This strategy encourages a more conservative policy because it aims to be resilient across different agent types. However, this broad applicability to a wide range of agent types sacrifices adaptability and responsiveness to boost/resist influence, as there is no expectation of benefit to customising responses to particular situations. Choosing a Non-TE agent for training the agents used in the human-agent experiments in Section \ref{subsec:human_agent_exp} was intended to prevent the dilution of communicative effects and to better represent the impact of influence modelling on human behaviour. Future research could focus on identifying user types to enable the selection of agents trained with corresponding user profiles. This approach could enhance the effectiveness of interactions by tailoring the agent's behaviour to specific user characteristics.

\begin{table*}[!htp]\centering
	\caption{Results training with Mixed-TE agents}\label{tab:mixed_result}
	\scriptsize
	\begin{tabular}{cccccccccc}\toprule
		\multirow{2}{*}{Experiment} &\multirow{2}{*}{Agent (TE type)} &\multicolumn{2}{c}{SRP (\%)} &\multicolumn{2}{c}{SRM (\%)} &\multicolumn{2}{c}{Averaged TE (bits)} &\multirow{2}{*}{CPS} \\\cmidrule{3-8}
		& &Mean &Std &Mean &Std &Mean &Std & \\\midrule
		\multirow{2}{*}{Random vs. Random} &P1 (Random) &80.00\% &- &20.00\% &- &- &- &\multirow{2}{*}{0.32} \\
		&P2 (Random) &80.00\% &- &20.00\% &- &- &- & \\
		\midrule
		\multirow{2}{*}{Mixed-TE vs. Non-TE} &P1 (Mixed-TE) &81.83\% &8.63\% &21.92\% &10.25\% &0.53 &0.38 &\multirow{2}{*}{0.36} \\
		&P2 (Non-TE) &86.12\% &7.99\% &26.25\% &10.25\% &0.49 &0.29 & \\
		\midrule
		\multirow{2}{*}{Mixed-TE vs. Pos-TE} &P1 (Mixed-TE) &\textbf{87.60}\% &7.48\% &\textbf{30.70}\% &10.21\% &0.78 &0.42 &\multirow{2}{*}{\textbf{0.42}} \\
		&P2 (Pos-TE) &87.51\% &7.64\% &30.64\% &10.66\% &1.26 &0.26 & \\
		\midrule
		\multirow{2}{*}{Mixed-TE vs. Neg-TE} &P1 (Mixed-TE) &80.21\% &9.02\% &20.91\% &9.61\% &0.29 &0.25 &\multirow{2}{*}{0.33} \\
		&P2 (Neg-TE) &81.52\% &9.10\% &22.04\% &9.78\% &0.54 &0.23 & \\
		\bottomrule
	\end{tabular}
\end{table*}


\section{Entropy analysis}
\label{apx:entropy_analysis}

\begin{figure*}[ht]
	\centerline{\includegraphics[width=1\linewidth]{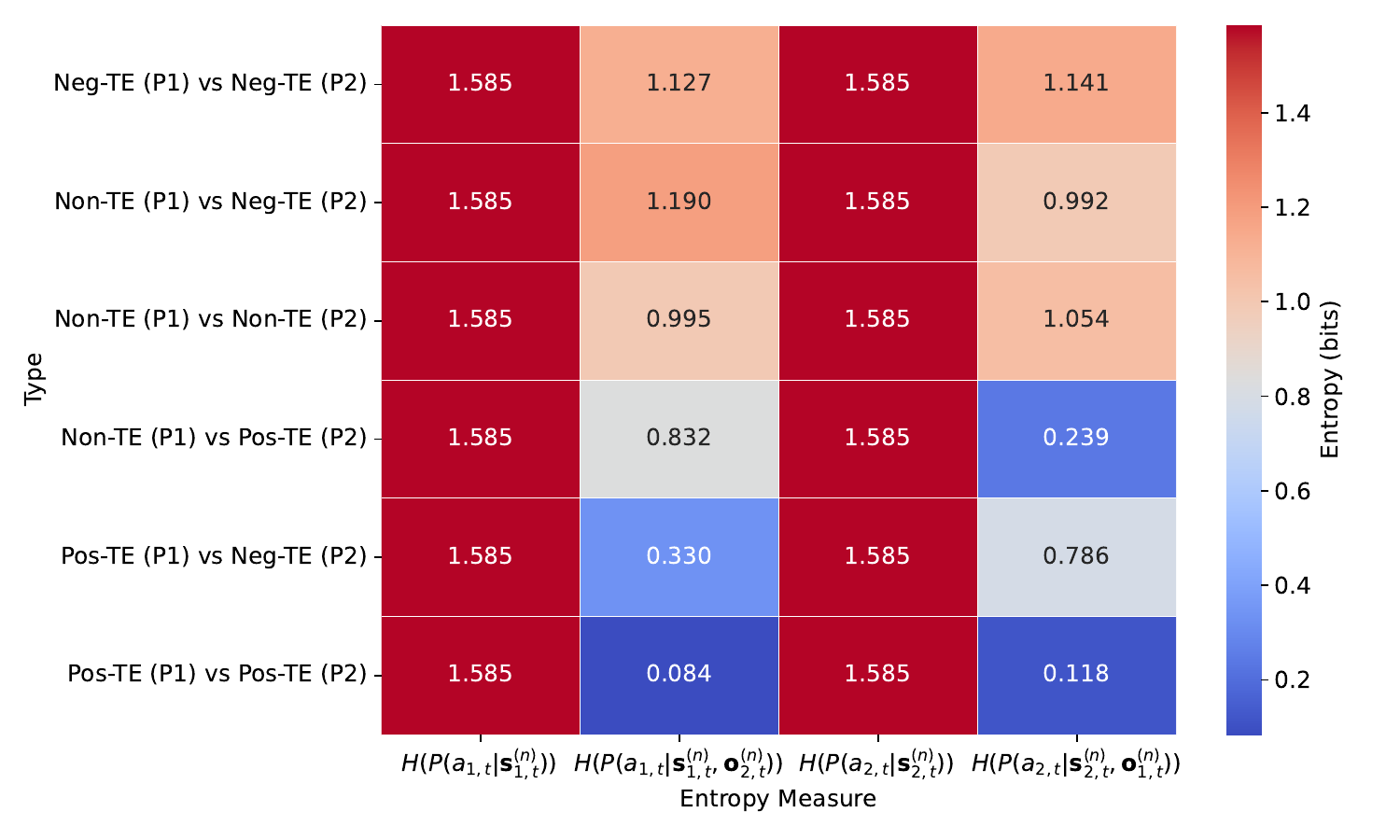}}
	\caption{Heat map of entropy measures. The first two columns are entropies for $P1$, and the last two columns are entropies for $P2$.}
	\label{fig:entropy_analysis}
\end{figure*}

We conducted further analysis to investigate the effect of influence modulation using TE by decomposing the TE into two entropy terms, as shown in Equation \eqref{get_te}. We ran the experiments with trained agents in Section \ref{subsec:simulation} for additional episodes and recorded their entropy measurements calculated according to Equation \eqref{comm_Q_values} and \eqref{nocomm_entropy}. Figure \ref{fig:entropy_analysis} presents the results in the form of a heat map. We observe that promoting influence (i.e., increasing TE) decreases the entropy of the actions with observations of the other agent, indicating that the ego-agent is more certain about their actions and receives more influence from the other agent. Conversely, resisting influence increases this entropy, which means the ego-agent cares less about the other agent. This finding supports our hypothesis in Section \ref{subsec:transfer_entropy}. Notice that the entropy, representing the uncertainty, of the action models without considering influence from others is always at its maximum (i.e., 1.585). This is due to the nature of the corridor dilemma task, where an agent cannot determine effective actions without observing the other agent. Nevertheless, the plotted averaged entropy measures provide a broad perspective rather than a detailed policy breakdown. It's essential to understand that manipulating TE isn't the same as solely tweaking entropy when considering observations of the other participants (i.e., $H(P(a_{1,t} | \mathbf{s}_{1,t}^{(n)}, \mathbf{o}_{2,t}^{(n)}))$ and $H(P(a_{2,t} | \mathbf{s}_{2,t}^{(n)}, \mathbf{o}_{1,t}^{(n)}))$). We verified this by repeating the same experiment with rewards solely aiming to decrease the entropy considering observations of the other participants. In this case the reward functions are written as:

\begin{equation}
	\label{entropy_ablation_reward}
	\begin{aligned}
		Reward_{a_{1,t}} &= -\phi H(P(a_{1,t} | \mathbf{s}_{1,t}^{(n)}, \mathbf{o}_{2,t}^{(n)})) + r,\\
		Reward_{a_{2,t}} &= -\phi H(P(a_{2,t} | \mathbf{s}_{2,t}^{(n)}, \mathbf{o}_{1,t}^{(n)})) + r.
	\end{aligned}
\end{equation}

We use the same settings as described in Appendix \ref{apx:simulation_details} for the experiments. The results are detailed in Table \ref{tab:entropy_albation_cc} and \ref{tab:entropy_ablation}. Comparing these results with those in Table \ref{tab:cc} and \ref{tab:additional}, it is evident that training with entropy considering observations from other participants has a negligible effect, further demonstrating the efficacy of the proposed framework.

\begin{table*}[!htp]\centering
	\caption{Results of training with entropy considering observations from other participants. The table shows SRCP and SRCL}\label{tab:entropy_albation_cc}
	\small
	\begin{tabular}{ccccccc}\toprule
		\multirow{2}{*}{Experiment} &\multirow{2}{*}{Agent (Entropy type)} &\multicolumn{2}{c}{SRCP (\%)} &\multicolumn{2}{c}{SRCL (\%)} \\\cmidrule{3-6}
		& &Mean &Std &Mean &Std \\\midrule
		\multirow{2}{*}{Random vs. Random} &P1 (Random) &50.00\% &- &\multirow{2}{*}{50.00\%} &\multirow{2}{*}{-} \\
		&P2 (Random) &50.00\% &- & & \\
		\midrule
		\multirow{2}{*}{Non-H vs. Non-H} &P1 (Non-H) &46.67\% &6.54\% &\multirow{2}{*}{64.85\%} &\multirow{2}{*}{17.96\%} \\
		&P2 (Non-H) &53.33\% &6.54\% & & \\
		\midrule
		\multirow{2}{*}{Non-H vs. Pos-H} &P1 (Non-H) &\textbf{59.25}\% &15.34\% &\multirow{2}{*}{53.77\%} &\multirow{2}{*}{12.14\%} \\
		&P2 (Pos-H) &40.75\% &15.34\% & & \\
		\midrule
		\multirow{2}{*}{Non-H vs. Neg-H} &P1 (Non-H) &55.30\% &13.93\% &\multirow{2}{*}{59.26\%} &\multirow{2}{*}{5.24\%} \\
		&P2 (Neg-H) &44.70\% &13.93\% & & \\
		\midrule
		\multirow{2}{*}{Pos-H vs. Pos-H} &P1 (Pos-H) &48.28\% &1.49\% &\multirow{2}{*}{\textbf{67.09}\%} &\multirow{2}{*}{13.29\%} \\
		&P2 (Pos-H) &51.72\% &1.49\% & & \\
		\midrule
		\multirow{2}{*}{Pos-H vs. Neg-H} &P1 (Pos-H) &54.65\% &1.43\% &\multirow{2}{*}{58.25\%} &\multirow{2}{*}{9.84\%} \\
		&P2 (Neg-H) &45.35\% &1.43\% & & \\
		\midrule
		\multirow{2}{*}{Neg-H vs. Neg-H} &P1 (Neg-H) &43.69\% &12.74\% &\multirow{2}{*}{50.00\%} &\multirow{2}{*}{13.61\%} \\
		&P2 (Neg-H) &56.31\% &12.74\% & & \\
		\bottomrule
	\end{tabular}
\end{table*}

\begin{table*}[!htp]\centering
	\caption{Results of training with entropy considering observations from other participants. The table shows SRP, SRM, CPS and averaged entropy in bits, namely, $H(P(a_{1,t} | \mathbf{s}_{1,t}^{(n)}, \mathbf{o}_{2,t}^{(n)}))$ and $H(P(a_{2,t} | \mathbf{s}_{2,t}^{(n)}, \mathbf{o}_{1,t}^{(n)}))$.}\label{tab:entropy_ablation}
	\scriptsize
	\begin{tabular}{cccccccccc}\toprule
		\multirow{2}{*}{Experiment} &\multirow{2}{*}{Agent (Entropy type)} &\multicolumn{2}{c}{SRP (\%)} &\multicolumn{2}{c}{SRM (\%)} &\multicolumn{2}{c}{Averaged entropy (bits)} &\multirow{2}{*}{CPS} \\\cmidrule{3-8}
		& &Mean &Std &Mean &Std &Mean &Std & \\\midrule
		\multirow{2}{*}{Random vs. Random} &P1 (Random) &80.00\% &- &20.00\% &- &- &- &\multirow{2}{*}{0.32} \\
		&P2 (Random) &80.00\% &- &20.00\% &- &- &- & \\
		\midrule
		\multirow{2}{*}{Non-H vs. Non-H} &P1 (Non-H) &87.04\% &7.66\% &27.83\% &10.44\% &1.02 &0.3 &\multirow{2}{*}{\textbf{0.39}} \\
		&P2 (Non-H) &85.72\% &8.05\% &26.33\% &10.54\% &1.03 &0.3 & \\
		\midrule
		\multirow{2}{*}{Non-H vs. Pos-H} &P1 (Non-H) &88.64\% &8.17\% &\textbf{28.11}\% &10.83\% &1.07 &0.32 &\multirow{2}{*}{0.38} \\
		&P2 (Pos-H) &84.67\% &7.85\% &23.22\% &9.83\% &0.46 &0.22 & \\
		\midrule
		\multirow{2}{*}{Non-H vs. Neg-H} &P1 (Non-H) &\textbf{89.07}\% &7.09\% &27.50\% &10.61\% &1.01 &0.3 &\multirow{2}{*}{\textbf{0.39}} \\
		&P2 (Neg-H) &87.54\% &8.20\% &25.28\% &11.44\% &0.6 &0.2 & \\
		\midrule
		\multirow{2}{*}{Pos-H vs. Pos-H} &P1 (Pos-H) &82.84\% &9.07\% &27.29\% &10.46\% &0.51 &0.22 &\multirow{2}{*}{\textbf{0.39}} \\
		&P2 (Pos-H) &83.71\% &8.20\% &28.07\% &10.25\% &0.52 &0.22 & \\
		\midrule
		\multirow{2}{*}{Neg-H vs. Neg-H} &P1 (Neg-H) &88.55\% &6.39\% &27.86\% &10.72\% &0.58 &0.2 &\multirow{2}{*}{\textbf{0.39}} \\
		&P2 (Neg-H) &87.68\% &8.33\% &26.26\% &10.35\% &0.56 &0.2 & \\
		\midrule
		\multirow{2}{*}{Pos-H vs. Neg-H} &P1 (Pos-H) &84.97\% &8.38\% &25.02\% &11.08\% &0.46 &0.22 &\multirow{2}{*}{0.37} \\
		&P2 (Neg-H) &85.34\% &8.47\% &24.56\% &10.91\% &0.58 &0.2 & \\
		\bottomrule
	\end{tabular}
\end{table*}


\section{Pairwise contrasts in simulation}
\label{apx:pairwise_contrasts_in_sim}

The pairwise contrasts of opponent type within each collaboration condition from the GLMM of the simulation are shown in Table \ref{tab:sim-glmm}.

\begin{table}[ht]
	\centering
	\caption{Pairwise contrasts of opponent type within each collaboration condition from the GLMM of the simulation. Estimates are on the log-odds scale. Odds ratios (OR) are shown for interpretability. P-values are Tukey-adjusted for multiple comparisons.}
	\label{tab:sim-glmm}
	\begin{tabular}{cccccc}
		\toprule
		Condition                      & Contrast        & Estimate (log-odds) & z     & p (Tukey) & Odds Ratio (OR) \\ \midrule
		\multirow{3}{*}{Collaboration} & Neg-TE - Non-TE & -0.22 & -4.24 & \textbf{0.0001} & \textbf{0.80} \\
		& Neg-TE - Pos-TE & -0.82 & -14.62 & \textbf{<.0001} & \textbf{0.44} \\
		& Non-TE - Pos-TE & -0.60 & -10.53 & \textbf{<.0001} & \textbf{0.55} \\ \midrule
		\multirow{3}{*}{Competition}   & Neg-TE - Non-TE & -0.02 & -0.41 & 0.91 & 0.98 \\
		& Neg-TE - Pos-TE & -0.41 & -7.91 & \textbf{<.0001} & \textbf{0.66} \\
		& Non-TE - Pos-TE & -0.39 & -7.50 & \textbf{<.0001} & \textbf{0.68} \\ \bottomrule
	\end{tabular}
\end{table}


\section{TE vs. Vehicle Proximity in the Highway Task}
\label{apx:te-vs-dist}

\begin{figure}
	\centering
	\includegraphics[width=1\linewidth]{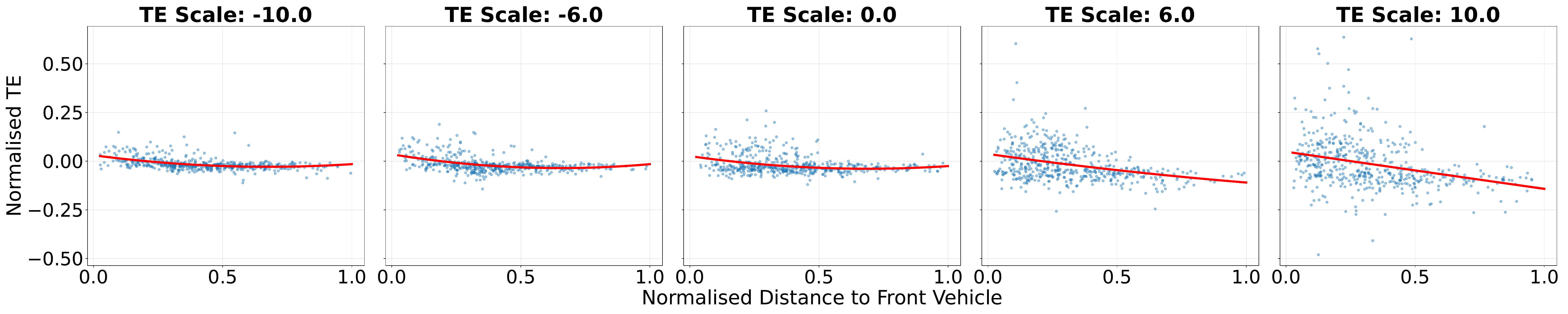}
	\caption{Normalised TE vs. Normalised Distance form the Ego-vehicle to the Other Vehicles with five different TE scales (i.e., $d$ in Equation \eqref{eq:highway-reward}.). Blue dots show 50\% of total samples (for ease of visualisation), and the red curves show quadric regressions.}
	\label{fig:te_vs_distance}
\end{figure}

We conducted an additional analysis to examine how TE varies with vehicle proximity. Figure \ref{fig:te_vs_distance} shows the normalised TE as a function of the normalised distance between the ego vehicle and other vehicles across five different TE scales (i.e., $d$ in Equation \eqref{eq:highway-reward}). The results indicate that TE increases when the inter-vehicle gap is short. Moreover, ego vehicles with positive TE tend to amplify this effect.
!


\end{document}